\documentclass[letterpaper, 11 pt, twocolumn,oneside]{IEEEtran}

\usepackage{graphicx}
\usepackage{flushend} 
\graphicspath{ {./images/} } 
\usepackage[boxruled,noline]{algorithm2e}
\usepackage{subcaption}
\usepackage{amsmath}
\usepackage{amssymb}
\usepackage{float}
\usepackage{amsfonts}
\usepackage{multirow}
\usepackage{tabularx}
\usepackage{algorithm2e}
\usepackage[colorlinks]{hyperref}
\usepackage[table]{xcolor} 
\usepackage{soul}
\usepackage[capitalise]{cleveref}

\begin{document}

%\title{ Reinforcement Learning Approach to Capacity Constrained Evacuation During an Active-Shooter Situation}
\title{An Optimized Evacuation Plan for an Active-Shooter Situation Constrained by Network Capacity}
\author{
\IEEEauthorblockN{Joseph Lavalle-Rivera\IEEEauthorrefmark{1}, Aniirudh Ramesh \IEEEauthorrefmark{2}, Subhadeep Chakraborty \IEEEauthorrefmark{3} }\\

\IEEEauthorblockA{\IEEEauthorrefmark{1}\emph{jlavall1@vols.utk.edu} \IEEEauthorrefmark{2}\emph{aramesh2@utk.edu}  \IEEEauthorrefmark{3}\emph{schakrab@utk.edu}} \\
\IEEEauthorblockA{Mechanical, Aerospace and Biomedical Engineering Department, University of Tennessee, Knoxville, 1512 Middle Dr, Knoxville, TN 37996, USA} 
}

\maketitle

\begin{abstract}

A total of more than $3400$ public shootings have occurred in the United States between 2016 and 2022. %\cite{Gunviolencearchive}. 
Among these, $25.1\%$ of them took place in an educational institution, $29.4\%$ at the workplace including office buildings, $19.6\%$ in retail store locations and $13.4\%$ in restaurants and bars. %\cite{peterson2021violence}. 
During these critical scenarios, making the right decisions while evacuating can make the difference between life and death. However, emergency evacuation is intensely stressful, which along with the lack of verifiable real-time information may lead to fatal incorrect decisions. To tackle this problem, we developed a multi-route routing optimization algorithm  that determines multiple optimal safe routes for each evacuee while accounting for available capacity along the route, thus reducing the threat of crowding and bottlenecking. Overall, our algorithm reduces the total casualties by $34.16\%$ and $53.3\%$, compared to our previous routing algorithm without capacity constraints and an expert advised routing strategy respectively. Further, our approach to reduce crowding resulted in an approximate $50\%$ reduction in occupancy in key bottlenecking nodes compared to both of the other evacuation algorithms.

\end{abstract}


%\IEEEpeerreviewmaketitle

\section{Introduction}

The topic of emergency evacuations boasts a rich literature; Liu et al. \cite{Liu2020} provide a comprehensive review with studies exploring a variety of approaches from agent-based models \cite{Lee2018agent,Hagh2021} and game theory \cite{Xiao2015,Vidal2002,Nowak1983} to statistical physics \cite{srinivasan2017pedestrian, karan2016dynamics}. Research topics range from human behavior (including vulnerable populations) effects \cite{Abdul2021,Duleb2019}, to modeling, simulation, optimization and guidance oriented studies \cite{Yin, Mira2021, Lin2020, Balb2020, Chou2019,  Tsai2021, Hong2018, Arteaga2023, srinivasan2018study, srinivasan2018Hazard}, and the use of immersive virtual reality-based experiments to study evacuation decision-making \cite{harris2023use}.%, and general dynamics of opinion evolution in society using computational models \cite{karan2017parametric,karan2017modeling, karan2016dynamics, karan2018effect} . 

Capacity-constrained evacuation routing optimization is more complex as it adds a temporal dimension to the graph environment \cite{Abus2020, Liu2022}. Optimization of egress paths in active- shooter scenarios is an even more complex task, not only due to the NP-hard nature of the optimization problem, but also because it must account for the dynamic and unpredictable nature of the threat as well as a multitude of psychological factors involved, such as social pressure, leader-follower and the ``familiar route is a safe route'' mentality. Gunn et al \cite{Gunn2017} is one of the very few works specifically on active-shooter evacuation where they divided evacuees into groups and used dynamic stochastic programming to find optimal routes for these groups. However, their work did not consider either the dynamic shooter movement, nor the capacity-constraint of the network. Lu et al. \cite{MRCCP} explored fast quasi-optimal capacity-constrained routing solutions by modeling capacity as a time series and used a heuristic algorithm called Multi-Route Capacity Constrained Planner (MRCCP). However, they did not consider the threat to be dynamic.  

% Routing optimization is a widely studied topic which find application in a wide range of industries from transpoC-CASTERStion and logistics to communication and networking. Simple forms of routing optimization such as the Dijkstra's algorithm do not account for the capacity constraints along the edges.  Consequently, such algorithms may not be suitable for more complex applications where crowding or bottlenecking need to be considered. These applications may need more sophisticated algorithms that satisfy capacity constraints along the edges, which may need multiple routes to avoid bottlenecks and maximize throughput.

% One such application is routing optimization for active shooter scenarios. Optimization of egress paths in active- shooter scenarios is a remarkably complex task, not only due to the NP-hard nature of the optimization problem but also because it must account for the dynamic moving threat and the capacity constraints along the routes in order to avoid bottlenecks and crowding. 

The active-shooter evacuation scenario presents different challenges that most other evacuations do not. Most evacuations require the evacuees to exit the area as quickly as possible, while an active shooter evacuation may require the evacuees to choose between hiding in relative safety or evacuating to an exit. 

Our previous work \cite{Lava2023} focused on establishing the viability of using Reinforcement Learning for routing optimization in active shooter scenarios. In that work, we showed that the optimized egress routes resulted in 56\% fewer casualties in a multi-parameter study. However, that study was focused on low occupancy cases that did not consider overcrowding and bottle-necking issues. 

In this work, we have developed a novel algorithm that computes and ranks multiple routes, accounts for the maximum and available capacities within rooms, hallways, and through doorways, and allows evacuees to avoid being in the shooter's line of sight by hiding or evading the shooter in order to maximize safety to address the issues of capacity constraints. We developed a discrete-event simulation using two topologically different building layouts to evaluate the effectiveness of our capacitated algorithm and compare it with other algorithms. Further, we have identified a method of ordering the routes that reduces the crowding and bottle-necking likelihood throughout the duration of an active shooter event. 

For the remainder of this paper, we summarize the current literature and other works in section II, describe the specifics of our methodology in section III, discuss the experimental design, the simulation environment, and the different algorithms we evaluated in section IV, discuss and compare the results from our simulations in section V, and, finally, identify future plans to improve the algorithm in section VI.

\section{Literature Survey}
Capacitated transportation networks have been a robust area of research within many different academic communities like operations research, supply chain optimization, theoretical computer science, etc. The approaches to solve these complex models are vast and varied within the current literature. The reasons that motivate planned and optimized movement are also varied, ranging from decreasing delays in general traffic flow to quick evacuation during natural disasters. Most of these events require potentially different approaches to satisfy domain-specific requirements. For example, modeling and optimizing the throughput across a capacitated transportation network to improve traffic flow can necessitate a very different approach to that of an emergency caused by a natural disaster. 

Emergency evacuations inside built environments are an interesting example of the capacitated transportation networks formulation.  A specific application within this subset is the evacuation during active shooter events.  Despite the large quantity of prior work in egress literature, a few key components of an active shooter evacuation makes it distinct from others - for one, the threat or danger in this situation is very mobile and moves in real time through the graph, which, along with the typical short duration of these events, may make it safer for evacuees to temporarily shelter in place or hide instead of trying to exit. Moreover, the danger posed by an active shooter is often spread within his line of sight and not solely around his physical proximity based upon the weapon in use. %The complexity of this problem may in part be a reason for the dearth of literature in egress for active-shooter scenarios.

Hong et al. modeled the evacuation of a building as a Multi-Objective Dynamic Route Network Planning (MODRNP) problem \cite{Hong2018}. Their approach uses iterative partition strategies to progressively identify shortest paths using Dijkstra's algorithm from the multiple sources to the multiple sinks. Li et al. use two methods, a maximum flow model (MFM) and a minimum cost maximum flow model (MC-MFM), to identify evacuation paths from a single source to either a single sink or multiple sinks\cite{Li2014}. Their results show that their models are practical and feasible, but don't necessarily avoid danger areas and only look to evacuate as efficiently as possible. Kaveh Shahabi and John Wilson examined road networks during wild fire evacuations and used their algorithm, Capacity-Aware Shortest Path Evacuation Router (CAS-PER), to identify shortest paths based upon available capacity\cite{Shah2018}. Their emphasis was placed on the ability to adjust routes based upon road closures, however, they continued to identify routes as a static problem with an adjusted graph network. Liu et al. established routes for building evacuation for multiple sources and multiple sinks by iteratively reserving capacity along shortest paths until all evacuees had exited\cite{Liu2016}. 

Using centrality measures of a graph to assist in the development of evacuation routes is an interesting and growing area of interest. Marin Lujak and Stefano Giordani used two metrics, evacuation betweeness-centrality and evacuation centrality, to identify and present what they refer to as agile routes\cite{Luja2018}. They classify agile routes as ones that are similar in length, but often also include similar alternate routes from the same source to the same sink that avoid certain parts of the network viewed as dangerous. In parallel work, Tamakloe et al. used a vehicle evacuation algorithm based upon a link-based centrality measure to identify "agile paths"\cite{Tama2021}. They used real traffic data from Seoul, Korea to demonstrate that their approach using centrality measure outperformed simple shortest path algorithms during high volume evacuation periods. The characteristic of being able to avoid specific parts of a network is of interest, however, in an active shooter situation, such danger zones are dynamic and staying in place may provide better safety. 

Another alternate approach is that of iterative path assignment heuristics. Qingsong Lue and Shashi Shekhar demonstrated that their heuristic, Multiple Route Capacity Constrained Planner (MRCCP), could achieve near optimal results by iteratively reserving capacity along the shortest routes\cite{MRCCP}. They primarily tested the planner on smaller networks to ensure near optimal results. To go a step further, Kim et al. produced another heuristic called the Intelligent Load Reduction heuristic that scales to larger networks that capacity constrained planners would struggle with\cite{Kim2007}. They are able to improve on computation time compared to the CCRPs by focusing on bottlenecks within the network and reducing the flow around them to account for the clogging of edges in the network. Neither of these methods account for a mobile danger as is the case in our work.

Tsai et al. examined a capacitated vehicle routing problem by comparing the use of a Deep Neural Network (DNN) based solution and that of a Sweep algorithm\cite{Tsai2021}. Using hurricane evacuation data from New Orleans, they hoped to show the DNN provided a vast improvement in the efficiency of routing when taking into account social distancing within the emergency vehicles. While efficiency did improve initially, the benefit provided by the DNN diminished greatly as the capacity of the emergency vehicle approached the average number of people per household being evacuated. The social distancing requirement increased the complexity of the routing and evacuation such that the use of smaller vehicles, which make up the vast majority of emergency transportation, almost completely removed the improvements of the DNN.

Acknowledging the risk or danger associated with a specific route or series of routes during evacuations is another aspect of pedestrian emergency evacuations that is well researched with many different approaches. Liu et al. examined evacuations aboard cruise ships that took into account route capacity and route safety\cite{Liu2022}. Interestingly, they did not avoid riskier paths, but instead gave evacuees on the riskier paths a higher priority to take the capacity along their route. The acceptance of route risk and optimization of speed along the route could potentially be applied to an active shooter situation as there is often a high risk-high reward case when trying to exit the building. Lee et al. developed a means of evaluating evacuation actions using AnyLogic models during active shooter events\cite{Lee2018}. They were able to examine the interactions of evacuation time, police response time, shots fired, etc. and evaluate the effectiveness that behavioral training, new technology, or architectural design of buildings had on the positive outcomes for the evacuees.

Arteaga et al. examined and scrutinized the effect that trained evacuation leaders had on positive evacuee outcomes during active shooter situations\cite{Arte2023}. They showed that the presence of trained evacuation leaders always improved the outcomes for the evacuees, regardless of the number of leaders present or their distribution. The leader-follower aspect of evacuation is of extreme interest to our application and believe it may greatly improve our modeling and planning in the future.

%The present work on pedestrian evacuation is large, but as stated earlier, the particular subset of active shooter evacuations is woefully under researched. 
In this paper, we propose a capacity-constrained framework based on Reinforcement Learning and a network optimization approach. Our approach of considering the active shooter as a dynamic threat with a dynamically changing risk metric and determining and distributing multiple evacuation routes that are ranked and allocated based upon danger metrics is novel and does not exist in the current literature to the best of our knowledge. 

\section{Methodology}

In \cite{Lava2023}, the structure of a Non-Homogeneous Semi-Markov Decision Process (NHSMDP) was established for optimizing the evacuation of evacuees from one and two floor buildings in the event of an active shooter situation. This paper established the use of a NHSDMP structure as a suitable framework for modeling the dynamics and interactions of the shooters and evacuees\cite{Geor2021}. Our previous work focused on the safe routing and movement of the evacuees throughout the buildings to avoid the threats posed by the shooter. We used a finite horizon reinforcement learning framework with time-expanded states to identify the optimal paths for the evacuees to avoid the shooter's line-of-sight. A wide range of simulation parameters including distribution of evacuees in the building, their movement speed relative to the shooter, location of first identification of the shooter, position update methods and frequencies were tested in which the NHSMDP method resulted in an overall 56\% reduction in casualties across two virtual buildings. 

One of the limitations of this work was the absence of capacity constraints along movement paths. As identified in the literature review above, capacity constraints in emergency evacuations are important to avoid bottle-necking and crowding.  Specifically in the context of active shooter scenarios, due to the dynamic nature of the active shooter threat, any bottle-necking near exits and narrow passageways would expose vulnerable people to serious harm. To address this important topic, using the same base framework as in \cite{Lava2023}, we introduce capacity constraints throughout the graph representations of the buildings on both nodes and edges. We deploy a Reinforcement Leaning- based algorithm on this graph that finds the optimal routes for the evacuees with a constraint on the maximum available route capacity and then iteratively finds the next optimal route until all evacuees have identified routes for safety.

\subsection{NHSMDP States}
The state in our NHSMDP structure is a three-tuple defined by $S = (node_{e} , node_{s} , t_{i})$ for $0\le i \le 300$ where $node_e$ and $node_s$ are the locations of the evacuee and the last detected location of the shooter respectively within the building, represented as a graph. The routes provided to the evacuees are computed based upon $node_e$ and a probabilistic risk measure of the likelihood of the evacuee being harmed by being in the shooter's line of sight, derived from $node_s$. Finally, the time at any instant, $t_i$, is the time from the start of an active shooter event ranging from $t_0$, the start of the event, to $t_{300}$, the end of an event measured in seconds. The finite time horizon was capped to $300$ seconds based on data from the Mother Jones active shooter database, which says the majority of active shooter events conclude within $5$ minutes\cite{Foll2022}. The total number of states depends on the graph environment and the number of discretized intervals within the $5$ minute interval. 

We examined the performance of multiple routing algorithms in two separate building environments that will be discussed in more detail later: an acyclic, single-level school, and a school building with two parallel hallways and multiple passages connecting them that is representative of a graph structure containing cycles. In the two environments we simulated, the Acyclic School had a state space of $907,500$ different states $(55 \times 55 \times 300)$ and the Cyclic School had $1,470,000$ states $(70 \times 70 \times 300)$. Movement from one state to another is associated with a sojourn time based upon the distance between the nodes. For instance, if $node_i$ is 5 seconds away from $node_j$, then leaving the state-tuple $(node_{i},node_{s},0)$ would result in potentially reaching the state-tuple $(node_{j},node_{s},5)$.

\subsection{NHSMDP Actions and Transition Probabilities}
\begin{figure}[h]
\centering
\includegraphics[scale = .75]{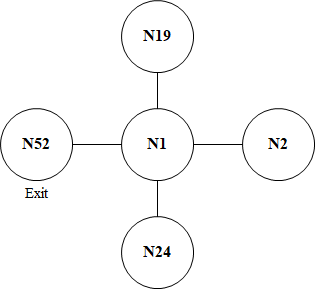}
\caption{Possible Actions in $node_1$}
\label{fig:posact}
\end{figure}

The possible actions that can be taken by an evacuee is based on the state that they are in and the available capacity at that time instant. Without considering capacity constraints, the possible actions in any given node are just the adjacent nodes to the agents current node. Consider a scenario where an evacuee is in $node_1$ in Figure \ref{fig:posact}. For an evacuee in $node_1$, the possible actions would be movement to $node_2$, $node_{19}$, $node_{24}$, or $node_{52}$ or they could choose to stay in their current node, $node_1$. 

However, when considering capacity constraints, all of these options may not be available when choosing an action. For instance, if $node_1$ and $node_{19}$ are separated by a door, (as shown in Fig. 4(a)), the throughput is dictated by the flowrate through that door. Studies have shown that about two people per second and four people per second are the closest approximations of throughput of a single-door and double door doorway, respectively \cite{Daam2010}. In the given state-tuple $(node_{1},node_{s},t_0)$, if there are two other evacuees moving through the doorway represented by $edge_{1-19}$ at $t_0$, then the available actions for the evacuee at $t_0$ are reduced down to $node_1$, $node_2$, $node_{24}$, or $node_{52}$. As the occupancy of nodes increases, especially around the exits, crowding and waiting can become a critical factor in the performance of any evacuation algorithm in such a severe situation. Hence, the challenge is to design an algorithm that minimizes crowding thereby improving throughput.

\begin{figure}[h]
\centering
\includegraphics[scale = 1]{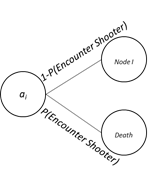}
\caption{Transition Probability for a Given Action $a_i$}
\label{fig:tranprob}
\end{figure}

The probability to transition from one state-tuple to another is also directly related to the probability of being within range of the shooter's weapon during the entirety of the transition and therefore does not remain constant as the shooter moves around the graph environment. When an evacuee takes an action, there are two possibilities: they reach the desired state-tuple or they come within range of the shooter's weapon as shown in Figure \ref{fig:tranprob}.

\begin{figure}[h]
\centering
\includegraphics[scale = .75]{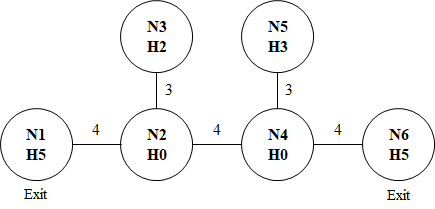}
\caption{ Probability Transition Example}
\label{fig:toyprob}
\end{figure}

The actual probability of encountering the shooter or being within range of his weapon is dependent upon his last known location and the passage of time. We explain this using an example seen in Figure \ref{fig:toyprob}. 

Each node in this example has a name, $N1$ for example, and a hardness, $H5$, that will be further explained in Section III C. Each edge has a value that indicates the sojourn time in between nodes. For this example, let us assume that we know the shooter is in $node_4$ at time $t_0$. Given the sojourn times and the available actions for the shooter, we compute the probability of the shooter's location over time as seen in Table \ref{table:slocprob}. We know where the shooter is at $t_0$ and the available actions are movement to $node_2$, $node_5$, or $node_6$ or remain in $node_4$. Without trying to presume any information about the shooter's intent, we assume an unbiased random walk model, i.e. he can go in any direction with equal probability. With that taken into account, at $t_1$, the shooter has an equal probability of being on $edge_{2-4}$, $edge_{4-5}$, $edge_{4-6}$, or in $node_{4}$ if he stayed in place. We also assume that once he has made a decision, i.e. selected a target node, he does not make a new decision until he reaches his target. His possible location is propagated throughout the graph and through time as seen in Table \ref{table:slocprob}.

Once the shooter's probability of visiting a specific node at a given time $t_i$ is rolled out over the planning time horizon, a risk metric informed by an estimate of the shooter's line-of-sight is calculated. The line of site from each node is pre-computed from the geometric layout of the building. Experts have suggested that the range of specific weapons that the shooter is carrying is less important than being in the line of sight of said weapon i.e. the threat from a rifle accurate out to 200m is the same as the threat from a handgun accurate out to 25m if the victim is in an enclosed hallway at a distance of 50m. The most important aspect in such a scenario is the need to stay out of line of sight. We approached this aspect of harm by smearing the maximum probability of a given node across all the nodes that are in its line of sight at any give time $t_i$.

\begin{table}[h]
\centering
\setlength{\tabcolsep}{5.2pt}
\begin{tabularx}{\columnwidth}{ |m{1.08cm}|c c c c c c c c| } 
 \hline
 Time in Seconds & 0 & 1 & 2 & 3 & 4 & 5 & 6 & 7 \\
 \hline
 $N_1$ & 0.00 & 0.00 & 0.00 & 0.00 & 0.00 & 0.00 & 0.00 & 0.00 \\
 \hline
 $N_2$ & 0.00 & 0.00 & 0.00 & 0.00 & 0.25 & 0.13 & 0.05 & 0.02 \\ 
 \hline
 $N_3$ & 0.00 & 0.00 & 0.00 & 0.00 & 0.00 & 0.00 & 0.00 & 0.06 \\ 
 \hline
 $N_4$ & 1.00 & 0.25 & 0.06 & 0.02 & 0.00 & 0.00 & 0.13 & 0.13 \\ 
 \hline
 $N_5$ & 0.00 & 0.00 & 0.00 & 0.25 & 0.19 & 0.11 & 0.06 & 0.03 \\ 
 \hline
 $N_6$ & 0.00 & 0.00 & 0.00 & 0.00 & 0.25 & 0.19 & 0.11 & 0.06 \\ 
 \hline
 $E_{1\rightarrow2}$ & 0.00 & 0.00 & 0.00 & 0.00 & 0.00 & 0.06 & 0.09 & 0.11 \\ 
 \hline
 $E_{2\rightarrow3}$ & 0.00 & 0.00 & 0.00 & 0.00 & 0.00 & 0.06 & 0.09 & 0.04 \\ 
 \hline
 $E_{2\rightarrow4}$ & 0.00 & 0.25 & 0.31 & 0.33 & 0.09 & 0.09 & 0.10 & 0.14 \\ 
 \hline
 $E_{4\rightarrow5}$ & 0.00 & 0.25 & 0.31 & 0.08 & 0.15 & 0.22 & 0.15 & 0.12 \\ 
 \hline
 $E_{4\rightarrow6}$ & 0.00 & 0.25 & 0.31 & 0.33 & 0.08 & 0.15 & 0.22 & 0.31 \\ 
 \hline
\end{tabularx}
\caption{Probable location of shooter at time=t.}
\label{table:slocprob}
\end{table}

\begin{table}
\centering
\setlength{\tabcolsep}{5.2pt}
\begin{tabularx}{\columnwidth}{ |m{1.08cm}|c c c c c c c c| } 
 \hline
 Time in Seconds & 0 & 1 & 2 & 3 & 4 & 5 & 6 & 7 \\
 \hline
 $N_1$ & 0.00 & 0.00 & 0.00 & 0.00 & 0.25 & 0.13 & 0.09 & 0.11 \\
 \hline
 $N_2$ & 1.00 & 0.25 & 0.31 & 0.33 & 0.25 & 0.13 & 0.13 & 0.14 \\ 
 \hline
 $N_3$ & 0.00 & 0.00 & 0.00 & 0.00 & 0.00 & 0.06 & 0.09 & 0.06 \\ 
 \hline
 $N_4$ & 1.00 & 0.25 & 0.31 & 0.33 & 0.25 & 0.22 & 0.22 & 0.31 \\ 
 \hline
 $N_5$ & 0.00 & 0.25 & 0.31 & 0.25 & 0.19 & 0.22 & 0.15 & 0.12 \\ 
 \hline
 $N_6$ & 1.00 & 0.25 & 0.31 & 0.33 & 0.25 & 0.19 & 0.22 & 0.31 \\ 
 \hline
\end{tabularx}
\caption{Probability of Harm by Shooter at time=t.}
\label{table:harmprob}
\end{table}

As can be seen in Table \ref{table:harmprob}, the shooter has line of sight on nodes $node_2$, $node_4$, and $node_6$. These nodes have the probability of encountering the shooter, physically or within the line of sight and range of the weapon, smeared up to the greatest value among the nodes that are currently within his line of sight. The process continues as location probability propagated through time as above. The harm array in Table \ref{table:harmprob} shows the difference between physically encountering the shooter as in Table \ref{table:slocprob} and simply being within his range.

\subsection{NHSMDP Reward Structure}

%The reward structure for this work is a modification from our previous published work. The previous reward function provided very small positive or negative rewards based upon the slight incremental improvement, or lack thereof, when moving from one node to another. While the small rewards worked adequately when capacity was not an issue, they created a lack of incentive to act in a predictable way. We designed the reward function, as seen below, to be more in line with standard reinforcement learning practices and to encourage more predictable actions.

\begin{algorithm}
    \SetKwFunction{Reward}{Reward}
    \SetKwInOut{KwIn}{Input}
    \SetKwInOut{KwOut}{Output}
    Moving from $node_i$ to $node_j$\;
    \uIf{$node_j$ = $death$}{
    reward = $-reward_{max}$ \;}
    \uElseIf{$node_j$ = $exit$}{
    reward = $reward_{max}$ \;}
    \uElseIf{$hardness_j$ - $hardness_i$ $\geq 0$}{
    reward = $1$ \;}
    \uElseIf{$exittime_j - exittime_i$ $<0$}{
    reward = $1$ \;}
    \Else{reward = $-1$ \;}
    \caption{Reward Structure for Actions}
\end{algorithm}

The $reward_{max}$ element in the reward function is the large reward received when an evacuee reaches a terminal state - a positive $reward_{max}$ by escaping, or a negative $reward_{max}$ by becoming a casualty due to proximity to the shooter's location. While the remainder of the reward function provides those interim rewards to help develop the actions and thus the routing, the $reward_{max}$ really provides the push and pull that drives the evacuees away from the shooter and towards an exit. The $hardness_i$ element represents that relative safety of $node_i$ - rooms with lockable doors have larger hardness index while nodes in exposed hallways have very small values. Using the previous example of moving from $node_1$ to $node_{19}$, the $hardness_{19}$ - $hardness_1\geq 0$ would result in a positive reward of $1$. Allowing the agent to receive the positive reward, even when there is no increase in hardness is vital as it allows an evacuee to remain hidden in a safer room without penalty. The $exittime_i$ is the time to travel to the nearest exit from $node_i$. Moving from $node_2$ to $node_1$ brings an evacuee closer to the exit at $node_{52}$ and hence would result in a positive reward of $1$. If none of these cases occur, the reward for the evacuee is $-1$ because they have not improved their situation by either going to, or remaining in, a safer node or by moving closer to an exit. We analyzed the affect that different values of $reward_{max}$ had on the performance of our evacuees through experimentation in the discrete event simulator. The results of the parameter exploration will be discussed further in the results section.

\subsection{Evacuation Algorithm - Capacitated Value Iteration}
%Our updated algorithm is an extension of the classic value iteration used in reinforcement learning. 
Finite horizon value iteration on the NHSMDP approach results in a single optimal policy for the evacuees to follow during the active shooter situation. However, the optimal policy, or route in our case, has capacity constraints that limit the number of evacuees who are able to enact that policy at a given time. As the routes approach the limits of capacity creating bottlenecks, additional routes need to be considered. %We developed a routing algorithm that determines multiple routes from each available node that accounts for all evacuees in each node that also intentionally reduces crowding and waiting in key nodes.

The Capacity-Constrained ASTERS (C-CASTERS) algorithm described in this paper is an extension of the NHSMDP based optimization, in which the possible actions available in each state are dynamically reduced as the available capacity for the action reaches zero. %We call our algorithm Real Time Adjacent ASTERS (C-CASTERS) and it is described briefly in the pseudo-code in Algorithm 2 below. The naming convention is demonstrative of its implementation within the python discrete event simulation. During the running of the simulation, we pause all movement of the shooter and evacuees and use the required inputs to process our routing algorithm. We then provide the new routes to the evacuees and resume the movements of the shooter and evacuees. The algorithm does not run in real-time during the simulation, but it is modeled as if it does and therefore is adjacent to real-time.

\begin{algorithm}
    \SetKwFunction{C-CASTERS}{C-CASTERS}
    \SetKwInOut{KwIn}{Input}
    \SetKwInOut{KwOut}{Output}
    \KwIn{\underline{Occupancy}: $E=$ $\sum e_i$ $\forall$ $node_i$ $\in$ $N$\\
        \underline{Shooter's Location}: $node_s$\\
        \underline{Node Order}: $OrNo$ = list of nodes $\in$ $N$\\
        ordered by proximity to an exit and\\
        then by proximity to $node_s$}
    \KwOut{A set of optimized routes $\forall$ $node_i$ $\in$ $ N$ that accounts for all evacuees, $E$, in $N$}

    \While {$E > 0$}{
        \For{$route_i$ $\in$ $route_{max}$}{
            \For{$node_j$ $\in$ $OrNo$}{
                Conduct Value Iteration\;
                Obtain Optimal Policy with \underline{$maxsend$}\;
                \For{$t \leftarrow 1$ \KwTo $30$}{
                Reserve \underline{$maxsend$} capacity along route through time\;
                Update transition probabilities affected by \underline{$maxsend$}\;
                }
            }
            Subtract \underline{$maxsend$} from $e_j$\;
        }
        Assign $route_i$ to evacuees\;
        Subtract $e_j$ from $E$\;
    }
\caption{C-CASTERS Pseudo Code}
\end{algorithm}

There are three inputs into the C-CASTERS algorithm: the current occupancy of the graphical environment, the shooter's known location, and an ordering of all of the nodes in the graphical environment, $OrNo$. Each node has a number of evacuees occupying it at the start of the algorithm, $e_i$, such that  $0\leq e_i \leq e_{max}$ where $e_{max}$ is the maximum occupancy of $node_i$ at any time. $E=\sum e_i$ is the total number of evacuees at a given time, which changes as evacuees escape or encounter the shooter. The second input is the shooter's last known location identified by an assumed monitoring system. The final input is the ordered list of nodes, $OrNo$, that establishes priority in two stages, first based on their proximity to an exit and then further ordered based on  proximity to the shooter with nodes closer to the shooter's position given priority. By ordering the nodes closest to the exits first, we are able to clear those nodes of evacuees and prevent mass crowding that would slow the overall evacuation process. Further, the secondary ordering based upon proximity to the shooter ensures that those closest to the shooter receive their routes with their capacity reserved before those that are farther away. In simulation, the crowding results of this ordering outperforms other approaches that do not take these factors into account.

The C-CASTERS algorithm initializes the transition probabilities based upon the shooter's known location and begins with the first set of routes, $route_1$. Starting at the first node in the ordered list of nodes with $e_i$ $\geq 0$, the algorithm conducts the value iteration and determines the optimal policy, in this case the optimal route, for $node_i$. The algorithm also determines the maximum amount of evacuees that can be sent along that optimal route, $maxsend$, which is determined by the smallest bottle neck along the optimal route. For instance in Figure \ref{fig:posact}, given that the available capacities on $edge_{2,1}$ and $edge_{1,52}$ are 20 and 4, respectively the $maxsend$ along the optimal exit path [$2,1,52$] from node $N2$ would be $4$. %The occupancy array that tracks the occupancy of all nodes and edges for the next 30 seconds is updated according the optimal route established by subtracting $maxsend$ from the available capacity along the entirety of the route to include $e_i$ at time $t=0$. 
Below is an example table showing how the capacity along the route [$2,1,52$] changes over time.

\begin{table}
\centering
\setlength{\tabcolsep}{6.2pt}
\begin{tabularx}{\columnwidth}{ |m{1.08cm}|c c c c c c c c| } 
 \hline
 Time in Seconds & 0 & 1 & 2 & 3 & 4 & 5 & 6 & 7 \\
 \hline
 $N_1$ & 20 & 20 & 20 & 20 & 16 & 20 & 20 & 20 \\
 \hline
 $N_2$ & 16 & 20 & 20 & 20 & 20 & 20 & 20 & 20 \\ 
 \hline
 $N_{52}$ & 999 & 999 & 999 & 999 & 999 & 999 & 995 & 999 \\ 
 \hline
 $E_{2\rightarrow1}$ & 20 & 16 & 16 & 16 & 20 & 20 & 20 & 20 \\ 
 \hline
 $E_{1\rightarrow52}$ & 4 & 4 & 4 & 4 & 4 & 0 & 4 & 4 \\ 
 \hline
\end{tabularx}
\caption{Capacity of Nodes and Edges Accounting for $maxsend$}
\label{table:changeocc}
\end{table}

For reference, the maximum capacity of $node_1$, $node_2$, $node_{52}$, $edge_{2\rightarrow1}$, and $edge_{1\rightarrow52}$ are 20, 20, 999, 20, and 4 respectively. The $maxsend$ value of 4 is reserved along the route through time changing the available capacity (Table \ref{table:changeocc}). The remaining available capacity can be used for other policies as they are determined. For instance, according to Table \ref{table:changeocc}, there is no more additional capacity on $edge_{1\rightarrow52}$ at time $t=5$ so any other evacuees in $node_1$ at $t=5$ would not be able to move to $node_{52}$ in the first second. The time to go through the door and move to $node_{52}$ from $node_1$ is $3$ seconds, but the capacity on that edge is used for the first second of the $3$ second travel time. As indicated earlier, the rate of travel through the doorway is two people per second, but if the travel takes longer than one second, then the edge needs to be opened for other evacuees even though the other travelers are still there. The assumption is that when moving through a doorway, the doorway is only blocked for the first second of travel and is then open for the remainder of the evacuee's travel time. The example below demonstrates how multiple routes from a single node provides option for staging the evacuation. For example, the third route computed from Node 19 suggests waiting for 3 seconds before proceeding to exit. These 3 seconds are vital for suppressing the crowding near the exit.

\begin{itemize}
    \item Route 1: [19, 1, 52]
    \item Route 2: [19, 19, 1, 52]
    \item Route 3: [19, 19, 19, 1, 52]
    \item Route 4: [19, 19, 19, 19, 1, 52]
    \item Route 5: [19, 19, 19, 19, 19, 1, 52]
\end{itemize} 

After reserving the capacity along the route, the algorithm would then move onto the next node in the ordered list and repeat the value iteration, optimal policy determination, and capacity reservation until each node with $e_i \geq 0$ has an optimal policy/route for the evacuees. These initial routes are assigned to the $maxsend$ number of evacuees and $E$, the sum of all evacuees without an optimal policy, is updated by subtracting those evacuees who have now been assigned a route. The algorithm restarts with a second round of routes for each $node$ that has $e_i \geq 0$ and continues until every nodes $e_i = 0$. Once all evacuees have been assigned capacity constrained routes, the simulation will resume with the new routes for the evacuees to follow.

\section{Experimental Design}

In order to conduct experiments and test the performance of our algorithm against that of other approaches we used a discrete event simulation package in python that provided us the flexibility to investigate all the important effects of capacity on the efficiency of the evacuation protocol. The experimental design section follows where the simulation environment and the actions of the agents and shooter are described along with an examination of the key parameters, and a comparison with alternate evacuation algorithms.

\subsection{Simulation Environment}

The simulation environment utilized was a discrete event simulation package in Python called SimPy. SimPy allows us to create many agents, both shooter and evacuees, with asynchronous updates, i.e. each take turns choosing actions and moving through the graph. The capacity constraints in nodes and along edges are accounted for by not allowing an agent to complete its chosen action if capacity occupation has been reached at any point along the chosen route. 

Each node and edge within the graphical environment is treated as a resource. Each of these resources has a finite capacity that cannot be overwhelmed due to the token taking mechanism that prevents too many users of a given resource. Each resource has a number of resource tokens, based upon their maximum capacity, which are requested and given to evacuees in a FIFO order. For example, an evacuee whose plan indicates movement from $node_1$ to $node_2$ would check to see if there is available capacity on $edge_{1 \rightarrow 2}$. If there was available capacity, they would turn in their resource token for $node_1$ in exchange for an $edge_{1 \rightarrow 2}$ token which they would hold until they reached $node_2$ where they would then turn in their $edge_{1 \rightarrow 2}$ token and request their $node_2$ token. If $edge_{1 \rightarrow 2}$ was already at full capacity, then they would remain in $node_1$ until there was available capacity on $edge_{1 \rightarrow 2}$ and follow the same procedure above. %This aspect of capacity is where the crowding metrics come from. If an algorithm does not account for capacity, then critical nodes adjacent to exits become very crowded as you have many evacuees seeking the same resources.

The simulations are run for $300$ seconds of simulation time with the python package checking each agents actions at every second and ensuring that everything occurs in the appropriate order. %The shooter acts first in every second of simulation to ensure that we were not inappropriately disadvantaging the shooter. 
Movements for each evacuee and the shooter are tracked for the entirety of the simulation. 

\subsection{Shooter Actions}

We assume that the shooter's plan or motivation is unknown prior to or during an occurrence. In the context of simulation, the shooter's actions are determined by first selecting a random target node, followed by the Dijkstra's shortest path from his current location to the target node. The shooter operates outside of the capacity constraints as anyone trying to share an edge or node with the shooter is automatically considered a casualty. The estimate for casualties is assumed to be deterministic based upon the proximity to the shooter and being within his line of sight. This is not realistic, but is an effective metric for consistent evaluation.

When the shooter reaches his target node, he waits for five seconds, chooses a new, never visited target node from the available remaining rooms, and begins moving. This process continues until all evacuees have either escaped, become casualties, or if the 300 seconds have passed. For each set of parameter combinations, we perturbed the initial conditions and used random seeds to run 20 simulations.

\subsection{Evacuee Actions}

The specific actions available to evacuees were discussed earlier in the value iteration preparation of the algorithm. We assume that the evacuees completely trust the algorithms and obey the instructions without hesitation or regard for their safety. This is an unrealistic but necessary assumption to avoid the variability associated with indecision, hesitation, or disobedience when comparing the performance of all the algorithms. Each evacuee is assigned a route at specific times, for example a route from $node_2$ could be [2, 1, 19, 19, 19, 19, 19, 19, 19, 19, 1, 52]. The evacuees follow the instructions until they receive a new route. In essence, the evacuee agents do not have any action choices as the algorithm acts as a single controller for all evacuee actions.

\subsection{Graph Environments}

We elected to use two different graphical environments for this work: a single level school that is acyclic and a topologically different school layout that is characterized by the presence of cycles and loops in the graph structure.

\begin{figure*}[h]
    \centering
    \begin{subfigure}[ht]{0.98\textwidth}
        \centering
        \includegraphics[height=5cm]{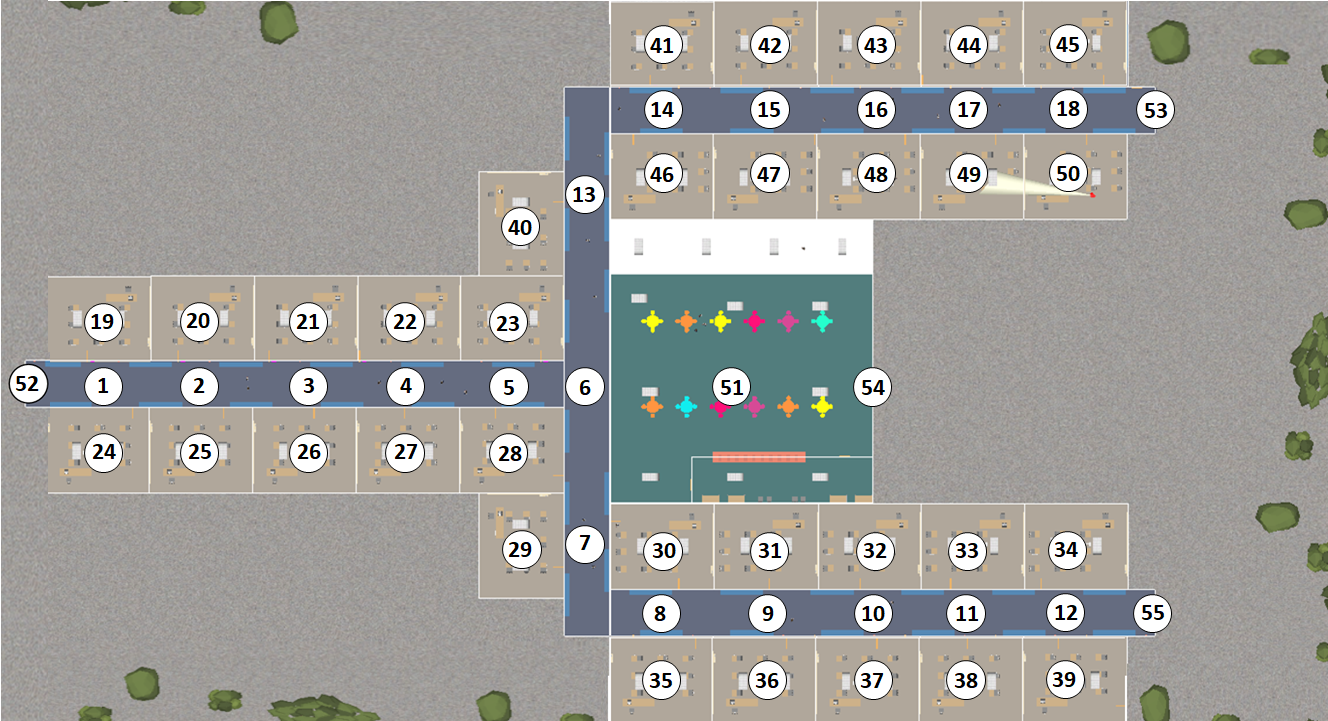}
        \caption{Acyclic School}
        \label{fig:uschool}
    \end{subfigure}
    \hfill
    \begin{subfigure}[ht]{0.98\textwidth}
        \centering
        \includegraphics[height=5cm]{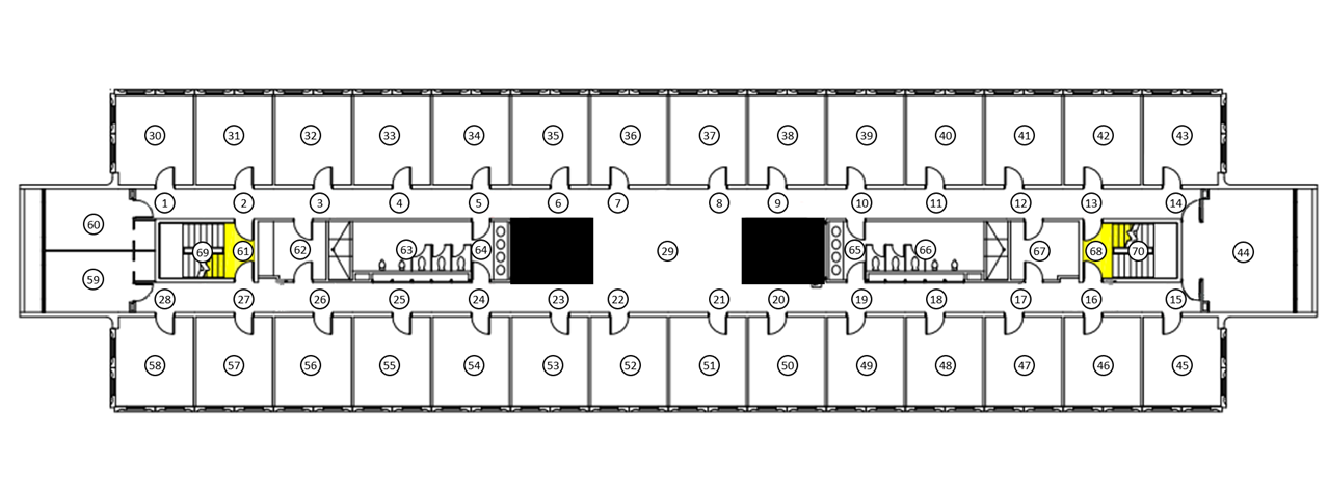}
        \caption{Cyclic School Floor Plan}
        \label{fig:wilson}
    \end{subfigure}
    \caption{Two Different Graphical Layouts for Simulation Comparison}
    \label{graphlayouts}
\end{figure*}

\subsubsection{Acyclic School}

The Acyclic School graphical environment consists of 55 total nodes as seen in Figure \ref{fig:uschool}. The node types are broken down as follows: Nodes 1 through 18 are hallway nodes, Nodes 19 through 51 are rooms nodes, and Nodes 52 through 55 are exit nodes. This graphical environment is useful as it does not provide cycles or loops where an agent can play cat and mouse with the shooter. It is a very straight forward layout that provides an excellent baseline understanding of the capabilities of the algorithms.

\subsubsection{Cyclic School}

The Cyclic School Floor is a graphical environment with two exits and 70 total nodes as seen in Figure \ref{fig:wilson}. The node type breakdown is as follows: Nodes 1 through 29 are considered hallway nodes, Nodes 30 through 68 are considered rooms nodes, and Nodes 69 and 70 are the exits. This graphical environment plays a key role in our analysis for two specific reasons. The first reason for its inclusion is the lines of sight down the hallway presenting dangerous possibilities when the shooter is in the hallway. The second, and more important reason, is that there are several connectors between the long hallways that allow both the shooter and evacuees to quickly move from one hallway to the other presenting the evacuees with a plethora of options, but also making the shooter's movements much more unpredictable and dangerous.

\subsection{Parameters and Simulation Inputs}

\subsubsection{Evacuee Distribution}\label{Sec:Evac_Dist}

How the evacuees are distributed at the onset of the event was the first of two important parameters studied. The evacuees were either distributed at the start of the simulation in only the rooms or uniformly distributed in both rooms and hallways. These two cases correspond to the typical bimodal distribution for movement within a school-day:

\begin{itemize}
    \item \textbf{Rooms and Hallways:} During recess or in between classes, there are students in both rooms and hallways milling about,
    \item \textbf{Rooms Only:} During class times all the students are typically in classrooms.
\end{itemize}  
Each spawn location, specific to the type of simulation was initialized with ten evacuees. For example, the Acyclic School graphical environment had 51 total non-exit nodes, Nodes 1 through 18 being hallway nodes and 19 through 51 rooms nodes. So, for the \textbf{rooms Only} case, there would be 330 evacuees with ten each initialized in Nodes 19 through 51. For the \textbf{Rooms and Hallways} simulation, we considered 510 evacuees with ten each in Nodes 1 through 51. 

\subsubsection{Shooter Spawn Location and Target Node}

The other primary parameter of interest and effect was the location at which the shooter was initialized, representing the first sighting of the shooter in a real-life scenario.  We chose the specific starting locations in each graph to capture a variety of situations to ensure that we adequately covered all situational possibilities. For the Acyclic School, we chose nine separate locations for the shooter to spawn: $node_2$, $node_6$, and $node_{11}$ for hallway nodes, $node_{20}$, $node_{38}$, and $node_{51}$ for room nodes, and $node_{52}$, $node_{54}$, and $node_55$ for exit nodes. For the Cyclic School, we chose eight spawn locations: $node_2$, $node_{16}$, and $node_{29}$ for hallway nodes, $node_{30}$, $node_{37}$, $node_{44}$, and $node_{59}$ for room nodes, and $node_{70}$ for the exit node.

Our assumption is that the primary target for the shooter will be a room within the building. With this in mind, each target node that the shooter chooses is randomly chosen from all the available unvisited rooms within the graphical environment. The proximity of the rooms to the shooter's current location increases the probability that they will be randomly chosen when a new target node is required. For instance, if the shooter's target node was 19 and he reached it, then the room nodes in the western hallway of the Acyclic School would have a higher probability of being chosen than those in the northern or southern wing of the school. The randomness of the chosen targets, both initially and throughout the simulation help to ensure that the movements of the shooter do not become predictable.

% \subsubsection{Shooter Location Update}

% During the simulation, the only information provided to the evacuees is the shooter's location, but the question was how often should that happen. Our previous work on this topic examined the affect that positional updates for the shooter had on the routing and safe movement of the evacuees within the graphical environments. In that work, we examined shooter location updates based upon both fixed-camera updates when the shooter was within sight and updates on set time-intervals. We examined positional updates at intervals of 1, 5, 10, 20, and 30 seconds. Through our experimentation we found that, while it was always better to have more information i.e. constant knowledge of the shooter's location, it was also unrealistic. On the other hand, 30 seconds was far too long of a time to go without any additional information provided to the evacuees. We settled on 20 second shooter position updates as the middle ground between too much information, constant or near-constant updates, with only marginal benefit and not enough in a realistic scenario, 30 second updates. During the simulation, the evacuees are provided with a new planned route based upon the shooter's updated location every 20 seconds.

\subsection{``Naive ASTERS Evacuation algorithm''}

The first algorithm that will be compared to the C-CASTERS algorithm is the naive ASTERS evacuation algorithm \cite{Lava2023}. The naive ASTERS algorithm was based on the assumption that there is always enough flow capacity through each path point that crowding and bottle necking is never an issue. The naive ASTERS algorithm conducts the value iteration in the NHSMDP framework and determines a single optimal route from each node and shooter location combination. %Every evacuee in a given node is assigned the same exact route. Using Figure \ref{fig:posact}, 10 evacuees in $node_{19}$ would all receive the same route of [19, 1, 52]. 
However, due to limitations in movement capacity through doorways, in the version of the problem under consideration, only two out of the ten evacuees would be able to execute that route immediately while the other eight would need to wait their turns (a few seconds) to execute it. This waiting adds up and cumulatively results in crowding at certain bottle-neck nodes.

\subsection{"Naturalistic Response"}

The Natural Response algorithm is based upon rules created by domain experts and the general guidance of the ``Run, Hide, Fight'' protocol. The Natural Response can be described in a series of steps:

\begin{itemize}
    \item Check the distance between evacuee position and shooter's last known position
    \item If the distance is $\geq D_{threshold}$ path lengths, use Dijkstra's to move to nearest exit away from the shooter
    \item If the distance is $< D_{threshold}$, either move to nearest room or remain in room
\end{itemize}

The Natural Response algorithm demonstrates the actions of an evacuee in a real world scenario in a logical, simplified manner: hide until the shooter is far enough away and then move to the exit. $D_{threshold}$ was selected to be $7$ based on the best performance achieved in a parameter sensitivity study. While this may keep evacuees safe, it also prevents many of the evacuee from escaping at the same time. Also, just like the naive ASTERS, this scheme suffers from overcrowding, since everyone in a given node will attempt to follow the same route and leading to bottlenecks near exits. This will be further examined in the results section.

\section{Results}

In this section, we will systematically compare the overall performance of C-CASTERS versus the two other algorithms, and then discuss parameters of interest and their effects on the performance of each of the algorithms. For the simulation results, 20 randomized runs of $300$ seconds were performed with each possible combination of parameter values to generate statistically significant data. For each of the results sections, we conducted simulations in both the Cyclic School and Acyclic School graphical environments, varied the starting locations with evacuees initially distributed in either rooms and halls or rooms only, and varied the initial spawn location for the shooter within the graphical environment with a randomized initial target node. 

\subsection{Effect of Reward Shaping}

%The reward function is characterized by several parameters, including $reward_{max}$ that represents the reward for escaping the situation or becoming a casualty. The value of $reward_{max}$ greatly affected the behaviors of the agents as you would expect. Our previous work focused on the trade-off of moving closer to an exit versus moving to a safer node which, in most cases, required different actions on the part of the agents. In our previous work where capacity was not a constraint, avoidance of the $-reward_{max}$ encouraged short term safety, however, as time went on, the pull of the $reward_{max}$ at an exit would move the agents out of their rooms and on a path to the exit. While this worked well and efficiently for a non-capacity constrained system, an agent who needs to worry about crowding and wait times may want to err of the side of caution when making these life and death choices. When dealing with many agents, we wanted to establish the premise that, because these events often end quickly, avoidance of the $-reward_{max}$, being in the proximity of the shooter, is just as valuable as achieving the $reward_{max}$, making it to an exit. So, we kept the values of $reward_{max}$ and $-reward_{max}$ the same in absolute value and concentrated on exploration of that value.

The reward function in our model is defined by several parameters, including $reward_{max}$, which represents the reward for escaping a dangerous situation, and $penalty_{max}$ representing encountering the shooter. The balance between $reward$ and $penalty$ dictates the trade-off between moving closer to an exit and moving to a safer node, often necessitating conflicting actions from the agents. In scenarios where capacity does not limit movement, agents prioritize short-term safety by avoiding the negative consequence associated with $penalty_{max}$. However, over time, the incentive of $reward_{max}$ at an exit provides agents with the impetus to leave their rooms and move towards the exit, leading to efficient outcomes in systems that are not capacity-constrained. In contrast, when crowding and wait times become concerns, agents need to prioritize more long-term caution in making efficient decisions. Given that such high-stakes events often unfold rapidly, we posited that avoiding $penalty$, i.e., remaining away from the immediate threat, is as critical as achieving $reward$ by reaching an exit. Therefore, we set the rewards and penalty values equal in absolute magnitude, that is, $penalty_{max}=-reward_{max}$, and focused our efforts on exploring the implications of this balance.

% We conducted experiments within our simulation environment with both assets and both evacuee distribution sets across values for $reward_{max}$ that ranged from $2$ to $25$. In practice, we found that $reward_{max}$ values that were $\leq 6$ or $\geq 14$ provided no change in performance or outcome from those two bounds, so the results of the experimentation only include values between $6$ and $14$. Initial exploratory analysis seemed to indicate the most effective value was in the vicinity of 10, so we concentrated our analysis around that value and extended towards our two bounds. For both graphical environments, we tested values of 6, 8, 9, 10, 11, 12, and 14. The results of the experimentation are below.

\begin{figure}[h]
\centering
\includegraphics[scale = .75]{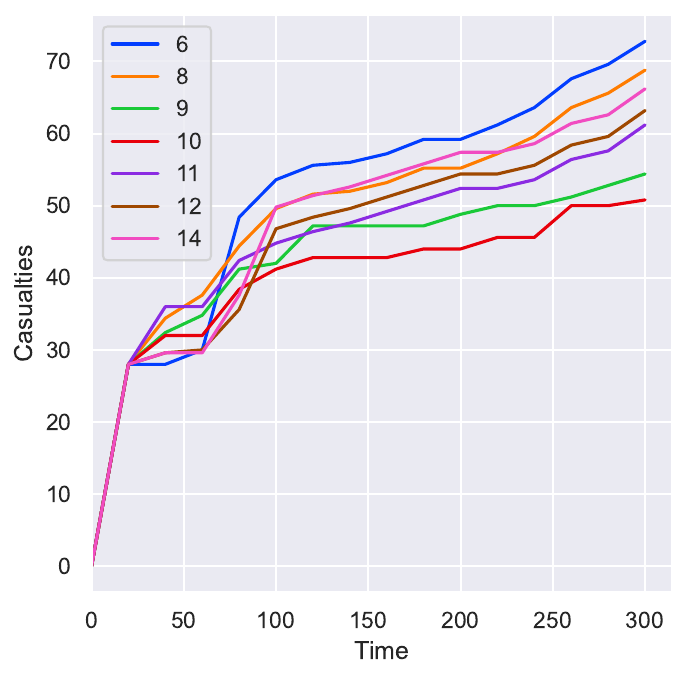}
\caption{Escape Reward Value Affect Over Time - Cyclic School Floor: Rooms and Halls Distribution}
\label{fig:WilsonRewa}
\end{figure}

\begin{figure}[t]
\centering
\includegraphics[scale = .75]{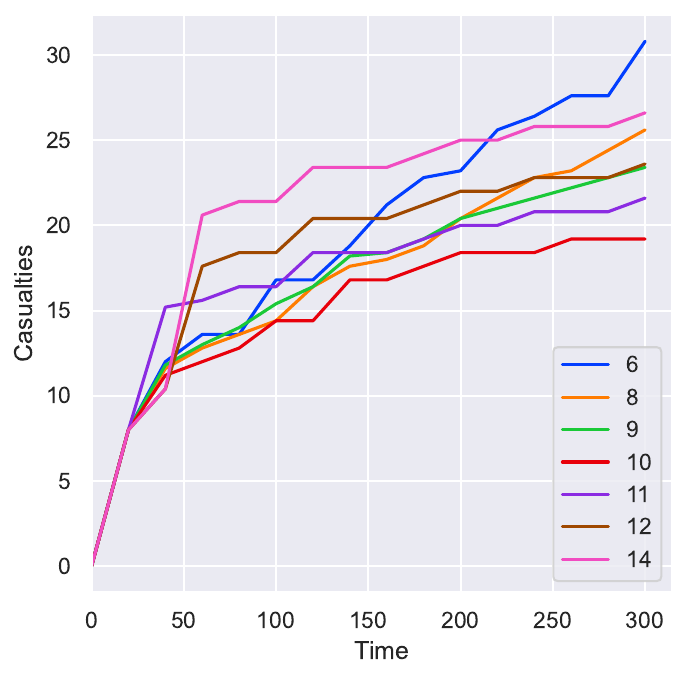}
\caption{Escape Reward Value Affect Over Time - Cyclic School Floor: Rooms Distribution}
\label{fig:WilsonRewb}
\end{figure}

\begin{figure}[t]
\centering
\includegraphics[scale = .75]{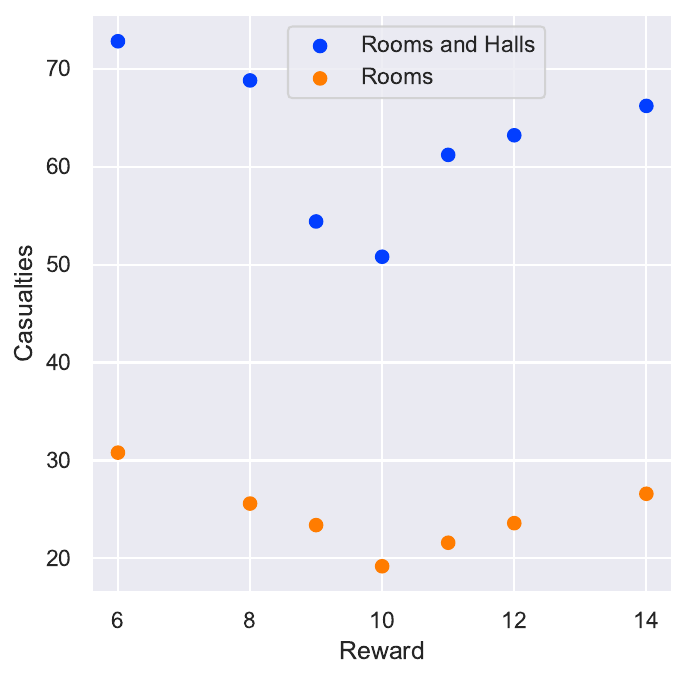}
\caption{Escape Reward Value Affect at 300 Secs - Cyclic School Floor}
\label{fig:WilsonRewc}
\end{figure}

\begin{figure}[t]
\centering
\includegraphics[scale = .75]{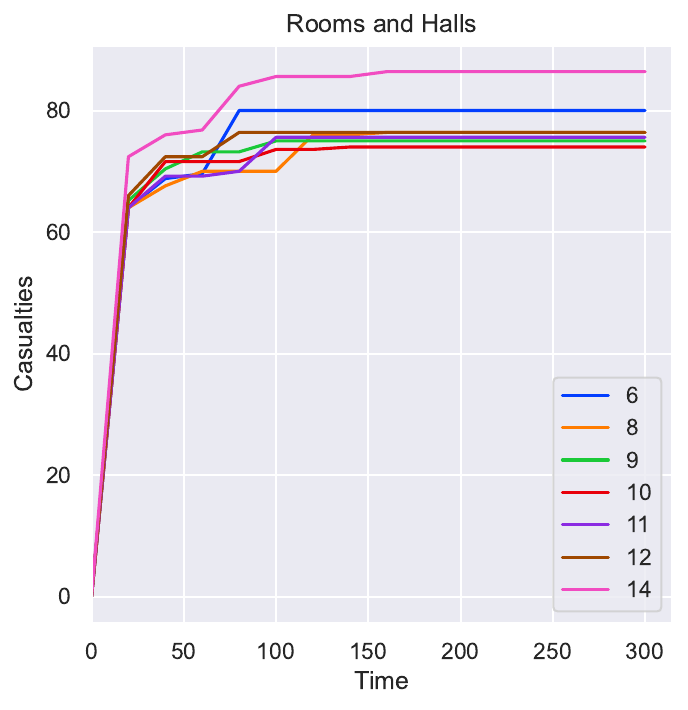}
\caption{Escape Reward Value Affect Over Time - Acyclic School: Rooms and Halls Distribution}
\label{fig:UnityRewa}
\end{figure}

\begin{figure}[t]
\centering
\includegraphics[scale = .75]{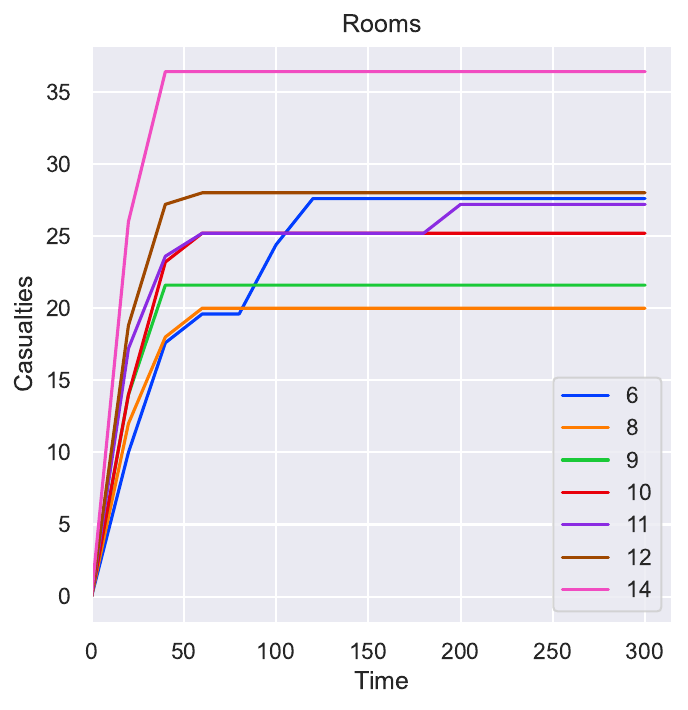}
\caption{Escape Reward Value Affect Over Time - Acyclic School: Rooms Distribution}
\label{fig:UnityRewb}
\end{figure}

\begin{figure}[h]
\centering
\includegraphics[scale = .75]{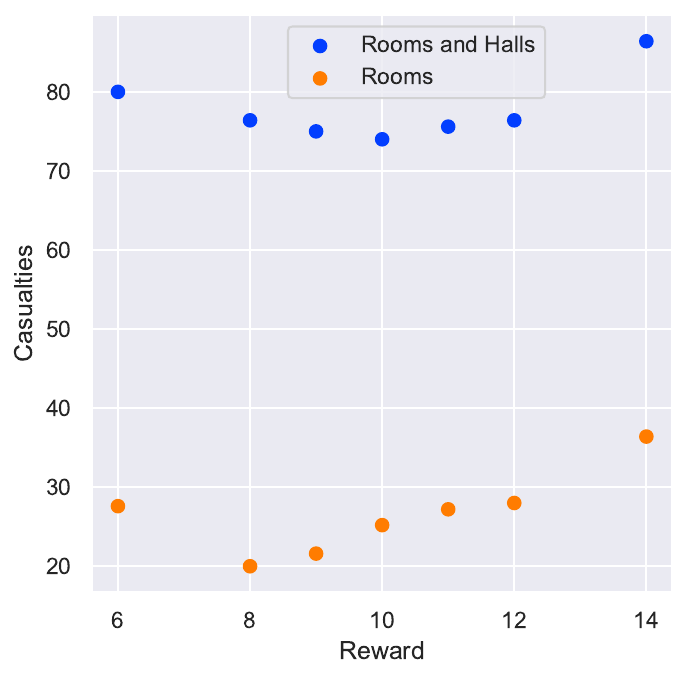}
\caption{Escape Reward Value Affect at 300 Secs - Acyclic School}
\label{fig:UnityRewc}
\end{figure}

As can be seen in Figures \ref{fig:WilsonRewa}, \ref{fig:WilsonRewb}, \ref{fig:WilsonRewc}, \ref{fig:UnityRewa}, \ref{fig:UnityRewb}, and \ref{fig:UnityRewc}, we examined the effect of varying values for the magnitude of $reward_{max}$ and $penalty_{max}$ on the number of casualties observed in the discrete-event simulation. Our goal was not only to establish the best $reward_{max}$ value for each of the possible combinations of parameters, but also to attempt to identify and examine common themes that occur naturally in the simulation results.

\textbf{Cyclic School Parametric Reward Shaping:} Figure \ref{fig:WilsonRewa} displays the cumulative averages of casualties of the $20$ runs over time. In this specific simulation, the evacuees were initially distributed in the rooms and halls in the Cyclic School Floor. %As can be seen in Figure \ref{fig:WilsonRewa}, all of the reward values have the exact same values for the first 20 seconds of simulation as the parameters have the same affect in the initial period of time. 
As can be seen in Figure \ref{fig:WilsonRewa}, in the initial 20 seconds, the number of casualties is completely independent of the evacuation strategy, since it can be argued that the initial loss of lives when a shooter starts firing cannot be prevented by optimized escape plans. However, after the first 20 seconds, an optimal risk-taking behavior (moderate values of $reward_{max}$) can be distinguished from the ``too safe'' and ``too risky'' plans. Interestingly, both the extreme low values, $\{6, 8\}$ and the extreme high values $\{12, 14\}$, provide the worst total casualty rates. Evacuees who are too risk averse, corresponding to low $reward_{max}$ values of $\{6, 8\}$, move less, even when it is likely safe to do so, which can result in greater total casualties. Aggressive risk-taking plans, corresponding to $reward_{max} \in \{12, 14\}$, incentivize escape attempts even when it is less safe to do so. This also results in greater number of casualties. It is interesting to note that the high-risk behavior results in more casualties earlier in the simulation, but then also leads to more escapes as the time moves beyond the first minute. This is in some way similar to the outcome achieved in the currently practiced ``run-hide-fight'' protocol, where "running" towards safety is prioritized whenever possible. This research suggests that a more balanced outlook may be safer with medium values of $reward_{max} \in \{9, 11\}$) achieving the optimal balance between staying safely away from the shooter while also attempting to move towards the exit when the conditions are more ideal.

Figure \ref{fig:WilsonRewb} displays the results when the evacuees are initially distributed in the relative safety of the rooms, in contrast to hallways, when the shooter is first located. Beyond an overall lowering of the number of casualties due to the initial protection offered by the rooms, the trends for this situation is very similar to that discussed before - an optimum level of risk tolerance ($reward_{max} = 10$) leads to the best outcome compared to both overly aggressive as well as  passive responses. % in the Cyclic Dorm Floor. It is important to note that, in most parameter combinations, the evacuees begin in safe positions due to their initial distribution in rooms only which explains the lower number of total casualties for each of the values of $reward_{max}$. As seen in Figure \ref{fig:WilsonRewa}, the extreme values leading to risk averse behaviors, $[6, 8]$, and risk taking behaviors, $[14, 12]$, result in the most casualties in the Cyclic Dorm Floor when the evacuees are initially distributed in rooms only. They also performed in a similar fashion in the first minute of the simulation with the risk averse behavior performing well and risk taking behavior performing poorly, and then switching after the first minute where the risk taking behavior performed better than the risk averse behavior. Also similar to previous results, the best performing values for $reward_{max}$ were $[10, 11, 9]$ which struck the optimal balance between safety and escape. The $reward_{max}$ value of $10$ was the best performing value for both evacuee distributions in the Cyclic Dorm Floor environment and hence will be the $reward_{max}$ value we use for the remainder of the simulations in that environment.

Figure \ref{fig:WilsonRewc} displays the average total casualties at the conclusion of the event ($5$ minute duration) for the different evacuee distributions in the Cyclic School Floor. Both distributions display a well-defined optimum around a $reward_{max}$ value of 10. The optimum balance between seeking safety and seeking escape, achieved with a $reward_{max}$ value of $10$ is specific to the specific graphical environment, but independent of the distribution of evacuuees, the initial location of the shooter in the graph and the randomness of the shooter's movement. %It is important to take note of the magnitude difference between the two evacuee distributions; evacuees being initially distributed in rooms only is a much safer circumstance that results in better safety outcomes. These results showed promise, but we also wanted to include the Acyclic School in the parameter exploration to determine if these values were universal or specific to different graphical environments.

\textbf{Acycilic School Parametric Reward Shaping:} Figure \ref{fig:UnityRewa} displays the results of the same parameter combination and runs for the Acyclic School as those conducted in the Cyclic School Floor environment. These graphical environments are topologically distinct and the results are indicative of that fact. The Acyclic School contains more exits than the cyclic school floor. Moreover, the lack of possible cycles connecting the hallways significantly changes the topology of the graph. Figure \ref{fig:UnityRewa} shows the results for the scenario when the evacuees are distributed in both the rooms and hallways. While the resulting performance of differing $reward_{max}$ values is very similar, it is important to note that the most risk taking behavior associated with a $reward_{max}$ of 14 performs the most poorly, especially early in the simulation demonstrating that breaking for an exit too soon is detrimental to survival. While other $reward_{max}$ values outperformed it within the first 100 seconds of simulation, a $reward_{max}$ value of 10 was again the best performing resulting in the least amount of casualties during the simulation. 

The results of distributing the evacuees in rooms only at the start of the simulation is displayed in Figure \ref{fig:UnityRewb}. The real differences between the $reward_{max}$ values occurs in the first minute of the simulation with each of the values producing very different results. Interestingly, Figure \ref{fig:UnityRewb} displays the only occurrence where 10 was not the best performing reward value among either graphical network or parameter combinations. The best performing values when the evacuees were distributed in rooms only were 8, 9, and 10 in order. While this result may cause one to question, in this case, whether smaller value of the reward outperformed the rest, you can see that a reward value of 6 was still poorer performing than all but 12 and 14. In a simpler environment such as this with fewer distinct pathing options, it is clear that being more risk averse is a better approach than being risk taking. In this environment, the balance between escaping and hiding is struck leaning towards hiding as it provides the best safety outcomes. 

Just like Figure \ref{fig:WilsonRewc}, Figure \ref{fig:UnityRewc} shows the average casualties at 300 seconds for each of the reward values and each of the evacuee distributions. The rooms and halls distribution still has the concave spread centered on 10, but, for the first time, the rooms only distribution showed a concavity centered on 8. This demonstrates to us that, when starting in a safe position in the simpler layout of the Acyclic School, it is more effective to err on the side of safety and hiding than moving to escape through an exit. While these results differ slightly from those of the Cyclic SChool Floor, they do support the contention that striking the balance between hiding and escape provides the most effective strategy even if the layout requires a behavior leaning more towards safety than escape.

Based upon these results, we elected to use the $reward_{max}$ value of 10 for all of the remaining parameter exploration runs contained in this work. There may be benefit in future work to adjust the reward function based upon certain assumptions and circumstances. For example, if we assume that an event is only going to last for 60 seconds, the $reward_{max}$ may potentially need to be smaller as the smaller values in general did well early in the parameter exploration and then worse as time went on. Perhaps a more complicated building may provide more hiding and movement opportunities than either of these graphical environment and would therefore encourage a larger $reward_{max}$ value. Tuning the $reward_{max}$ value to the building and the situation is the most likely way to improve the performance of the routing and the safety outcomes of the evacuees.

\subsection{Aggregated Performance}

\begin{table*}[t]
\makebox[\textwidth][c]{       %centering table
\resizebox{\textwidth}{!}{   %resize table
\begin{tabularx}{.855\paperwidth}{ |m{.75em} |m{1.5em}|c |c |c |c |c |c |c | } 
 \hline
 \multicolumn{3}{|c|}{Shooter Spawn} & \multicolumn{2}{|c|}{Exit} & \multicolumn{2}{|c|}{Hall} & \multicolumn{2}{|c|}{Room}\\
 \hline
 \multicolumn{3}{|c|}{Evacuee Distribution} & {Rooms} & {Rooms and Halls} & {Rooms} & {Rooms and Halls} & {Rooms} & {Rooms and Halls}\\
 \hline
 \multirow{9}{*}{\rotatebox[origin=c]{90}{\parbox{3.5cm}{Acyclic School}}} 
   & \multirow{3}{1.75cm}{\rotatebox[origin=c]{90}{\parbox{1.75cm}{\centering C-CASTERS}}} 
   & Casualties & 13.4(4.06\%) & \textbf{25.67(5.03\%)} & \textbf{20.07(6.08\%)} & \textbf{27.4(5.37\%)} & 23.6(7.15\%) & \textbf{35.07(6.88\%)}\\ [1.45ex] 
  \cline{3-9}
  &   & Escapes & 316.6(95.94\%) & \textbf{461(90.39\%)} & \textbf{309.93(93.92\%)} & \textbf{429.27(84.17\%)} & \textbf{306.4(92.85\%)} & \textbf{464.93(91.16\%)}\\[1.45ex]
  \cline{3-9}
  &   & LOS & \textbf{44.63} & 258.83 & \textbf{75.63} & \textbf{254.8} & \textbf{45.67} & \textbf{212.1}\\[1.45ex]
  \cline{2-9}
  & \multirow{3}{1.5cm}{\rotatebox[origin=c]{90}{\parbox{1.5cm}{Naive\\ ASTERS}}} 
  & Casualties & \textbf{8.95(2.71\%)} & 51.5(10.1\%) & 23.91(7.24\%) & 64.71(12.69\%) & \textbf{18.53(5.62\%)} & 93.33(18.3\%)\\[1.45ex]
  \cline{3-9}
  &   & Escapes & \textbf{317.59(96.24\%)} & 447.67(87.78\%) & 306.09(92.76\%) & 404.45(79.3\%) & 301.33(91.31\%) & 418.91(82.14\%)\\[1.45ex]
  \cline{3-9}
  &   & LOS & 45.03 & \textbf{217.72} & 99.08 & 290.69 & 67.59 & 217.72\\[1.45ex]
  \cline{2-9}
  & \multirow{3}{1.5cm}{\rotatebox[origin=c]{90}{\parbox{1.5cm}{Natural\\ Response}}} 
  & Casualties & 56.76(17.2\%) & 84.23(16.52\%) & 51.91(15.73\%) & 71.57(14.03\%) & 65.64(19.89\%) & 122.28(23.98\%)\\[1.45ex]
  \cline{3-9}
  &   & Escapes & 269.57(81.69\%) & 402.01(78.83\%) & 277.65(84.14\%) & 384.68(75.43\%) & 253.84(76.92\%) & 377.29(73.98\%)\\[1.45ex]
  \cline{3-9}
  &   & LOS & 405.97 & 789.83 & 225.97 & 452.69 & 307.28 & 542.01\\[1.45ex]
  \cline{2-9}
  \hline
  \multirow{9}{*}{\rotatebox[origin=c]{90}{\parbox{3.5cm}{Cyclic School}}} 
  & \multirow{3}{1.75cm}{\rotatebox[origin=c]{90}{\parbox{1.75cm}{\centering C-CASTERS}}} 
  & Casualties & \textbf{18.8(12.05\%)} & \textbf{39.6(14.56\%)} & \textbf{19.17(12.29\%)} & \textbf{25.73(9.46\%)} & \textbf{16.35(10.48\%)} & \textbf{31.43(11.55\%)}\\[1.45ex]
  \cline{3-9}
  &   & Escapes & 79.8(51.15\%) & 140.2(51.54\%) & 83.97(53.82\%) & 142.2(52.28\%) & 83.3(53.4\%) & 151.08(55.54\%)\\[1.45ex]
  \cline{3-9}
  &   & LOS & \textbf{23.4} & \textbf{100.8} & \textbf{33.77} & 127.9 & \textbf{27.5} & \textbf{55.5}\\[1.45ex]
  \cline{2-9}
  & \multirow{3}{1.5cm}{\rotatebox[origin=c]{90}{\parbox{1.5cm}{Naive\\ ASTERS}}} 
  & Casualties & 27.36(17.54\%) & 55.56(20.43\%) & 27.16(17.41\%) & 37.15(13.66\%) & 22.84(14.64\%) & 50.28(18.49\%)\\[1.45ex]
  \cline{3-9}
  &   & Escapes & 106.52(68.28\%) & 182.04(66.93\%) & 110.39(70.76\%) & 187.61(68.98\%) & 118.58(76.01\%) & 198.01(72.8\%)\\[1.45ex]
  \cline{3-9}
  &   & LOS & 102.64 & 246.08 & 78.53 & 259.35 & 61.79 & 226.86\\[1.45ex]
  \cline{2-9}
  & \multirow{3}{1.5cm}{\rotatebox[origin=c]{90}{\parbox{1.5cm}{Natural\\ Response}}} 
  & Casualties & 30.8(19.74\%) & 53.48(19.66\%) & 25.92(16.62\%) & 38.8(14.26\%) & 26.57(17.03\%) & 53.26(19.58\%)\\[1.45ex]
  \cline{3-9}
  &   & Escapes & \textbf{121.2(77.69\%)} & \textbf{218.52(80.34\%)} & \textbf{130.08(83.38\%)} & \textbf{221.87(81.57\%)} & \textbf{125.43(80.4\%)} & \textbf{218.74(80.42\%)}\\[1.45ex]
  \cline{3-9}
  &   & LOS & 429.76 & 726.6 & 605.35 & \textbf{1002.2} & 406.9 & 655.86\\[1.45ex]
  \cline{2-9}
  \hline
 \end{tabularx} }}
\caption{\centering Aggregated Tabular Simulation Results Across Assets and Algorithms: This table shows the performance of each algorithm for each combination of parameters for both the Acyclic School and Cyclic School assets. The values in bold represent the best performance amongst all the algorithms for each possible parameter combination.}
\label{table:4}
\end{table*}

The performance of the algorithms within the two graphical environments are affected by various factors which will be discussed in subsequent results sections. The analysis of the aggregated performance of the algorithms takes all of the various parameter combinations into account without isolating specific values or combinations. Table \ref{table:4} displays the overall results for each of the algorithms and shows the parametric combinations of features with algorithmic specific performance based upon the average number of casualties and escapes in raw numbers and by percent of the total evacuees as well as the total number of seconds all of the evacuees spent in the shooter's line of sight. The casualties entries are the average. The three primary metrics that we captured during simulation were the number of casualties, the number of escapees, and the time spent in the line of sight of the shooter, whether the evacuee was close enough to be harmed or not. Our objective was to minimize the number of casualties and the time in line of sight as law enforcement experts agree that minimizing it provides the highest probability of survival. For all of the remaining discussions, an evacuee was deterministically considered to be a casualty if they, at any instant of time, were within three path-lengths of the shooter and were within the shooter's line of sight. If they were in the shooter's line of sight but were farther than three path-lengths away from the shooter then they were considered to be in his line of sight only, but were free to continue their movement. Evacuees who escape are those that reach an exit node before being harmed by the shooter. We did not place significant emphasis on the escape metric because it does not necessarily equate to the best outcome. For example, an aggressive routing plan may allow 75\% of the evacuees to escape while leading the remaining 25\% to become casualties, while a less aggressive routing plan may only result in 60\% escapes but also reduce casualties to 10\% with the remaining 30\% remaining safely in the building and away from the shooter until the emergency has passed. %We contend that reducing the casualties and the time in line of sight is more important than simply forcing the maximum number of escapes.

\begin{figure}[t]
\centering
\includegraphics[width=.98\columnwidth]{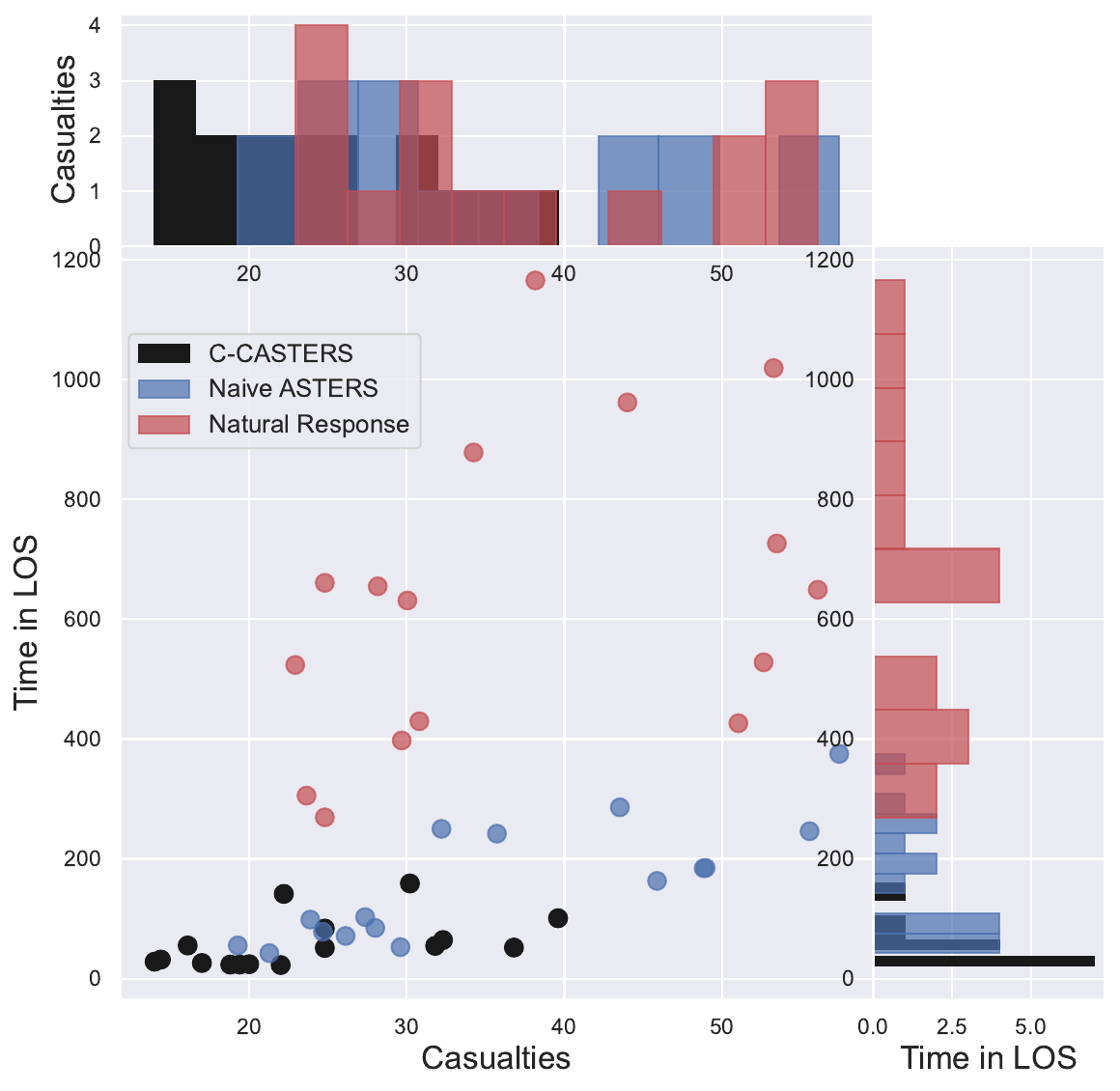}
\caption{Aggregated Evacuation Effectiveness - Cyclic School Floor}
\label{fig:WilsonHS}
\end{figure}

\textbf{Cyclic School Algorithm Performance:} Figure \ref{fig:WilsonHS} shows the combination scatter plot and histograms for the interactions between the number of casualties and the time spent in line of sight of the shooter for the Cyclic School Floor graphical environment. The casualty histogram displays how each of the three algorithms performed according to the casualties accumulated. The C-CASTERS algorithm outperforms the other two algorithms with an average of $24.01$ casualties, while Naive ASTERS averages $35.52$ casualties and Natural Response averages $37.36$ casualties. The line of sight metric is the cumulative time spent in the shooter's line of sight for all of the evacuees throughout the simulation. The time in line-of-sight histogram shows the same result with C-CASTERS outperforming the others with a cumulative average of $58.83$ seconds. Naive ASTERS comes in second with an average of $157.31$ seconds, and the Natural Response algorithm, averaging $639.38$ seconds is the worst performer. For the scatter plot, the best values are located in the lower-left and the worst are in the upper right quadrant. The plot demonstrates the way in which C-CASTERS outperforms the other two algorithms with consistently lower values for time in line of sight and fewer casualties and also lending support to the law enforcement point of view that decreased time in the shooter's line of sight results in fewer casualties.

\begin{figure}[t]
\centering
\includegraphics[width=.98\columnwidth]{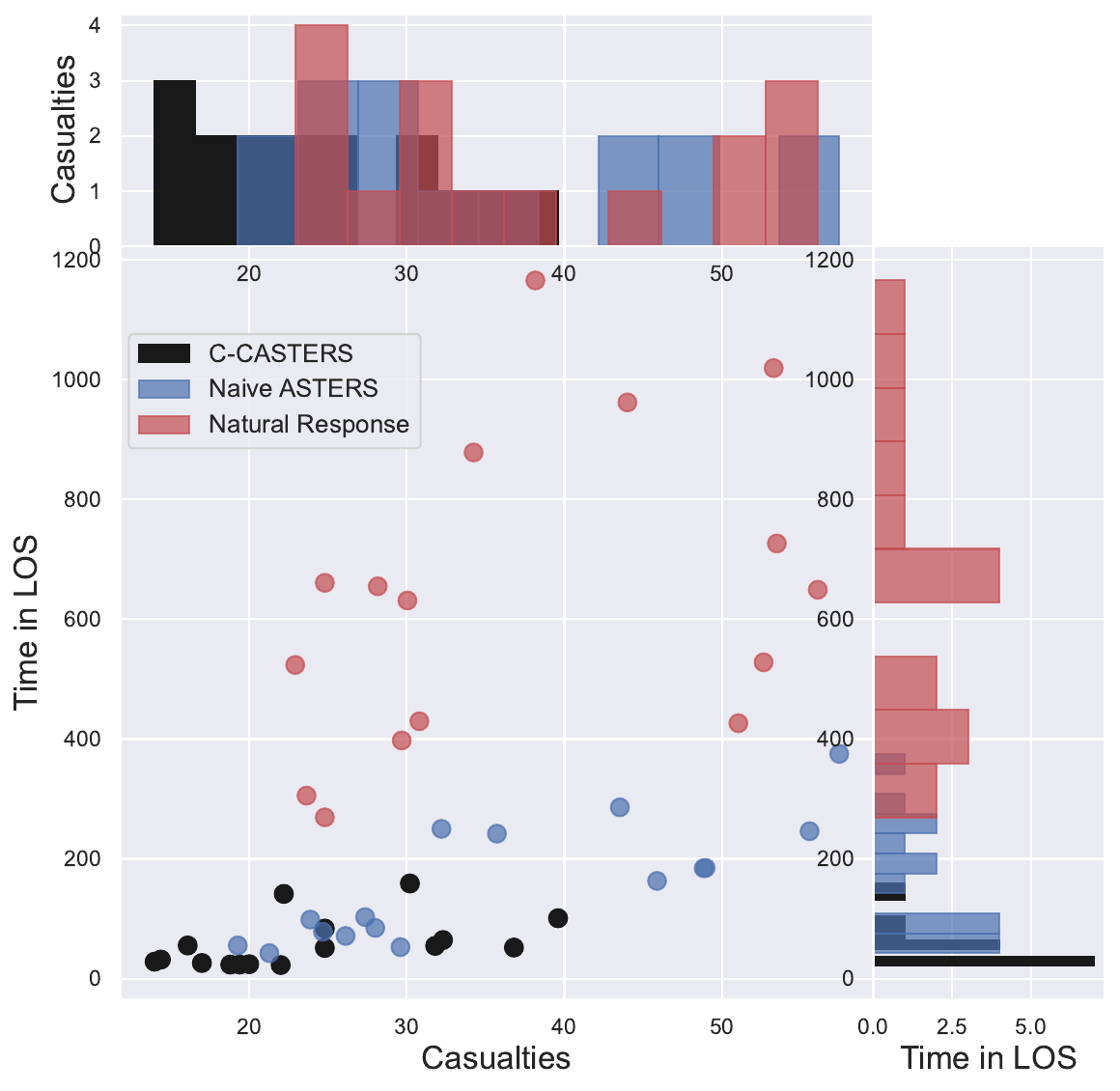}
\caption{Aggregated Evacuation Effectiveness - Acyclic School}
\label{fig:UnityHS}
\end{figure}

\textbf{Acyclic School Algorithm Performance:} The Acyclic School graphical environment results are displayed in the same type of figure in Figure \ref{fig:UnityHS}. Similar to the Cyclic School Floor environment, the C-CASTERS algorithm outperforms the other two with an average of $24.2$ casualties with Naive ASTERS having an average of $43.49$ casualties and the Natural Response having $75.40$ average casualties. The simpler layout of the Acyclic School broadens the differences between the algorithms and highlights the benefits of a capacity constrained approach. The C-CASTERS and Naive ASTERS were very close in terms the line-of-sight metric with C-CASTERS averaging $148.61$ seconds and Naive ASTERS averaging $156.30$ seconds. Natural Response was again severely outperformed averaging $453.96$ seconds. Similar to the Cyclic School Floor, Figure \ref{fig:UnityHS}'s scatter plot shows the concentration of the C-CASTERS in the lower left quadrant with the Naive ASTERS stretched a bit more along the casualties axis and the Natural Response having an even greater spread. Accounting and planning for capacity constraints greatly improved the safety outcomes for the evacuees in both graphical environments and provided promising insights for future development.

\subsection{Effect of Shooter's Start Location}

The location where the shooter is first detected and/or reveals himself by using his firearm, can have a drastic effect on the performance of any routing algorithm. We chose several locations in different regions of the graphical environments to ensure that we captured all possible situations of shooter movement through the graphs. For each of the 20 runs for a given scenario, the shooter was spawned in the same location but was given a randomly chosen target node that he moved toward, thereby creating randomness and difference between each run. For the sake of this analysis, we categorized the shooter's starting locations into one of three possibilities: Exits, Halls, or Rooms. 

\begin{figure}[t]
\centering
\includegraphics[width=.98\columnwidth]{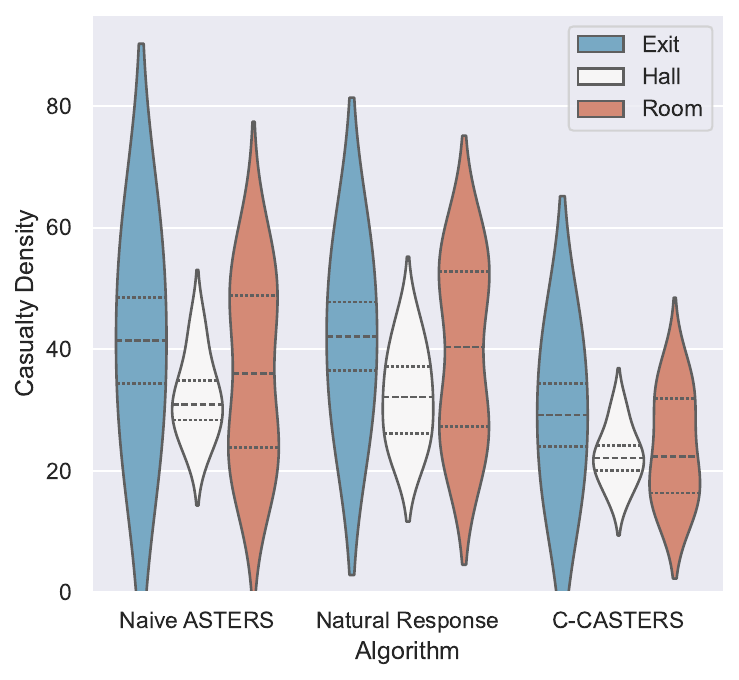}
\caption{Shooter's Spawn Location Effect - Cyclic School Floor}
\label{fig:WilsonSS}
\end{figure}

\textbf{Cyclic School Shooter's Start Location Effects:} Figure \ref{fig:WilsonSS} displays violin plots of the number of casualty for the different starting position categories of the shooter in the Cyclic School Floor graphical environment. Each of the element of the violin plots in essence is a vertical, smoothed histogram that identifies the mean and the quartiles. Longer, stretched ``violins'' are indicative of a large spread with greater variance of the given data while  shorter, fatter ``violins'' are indicative of of densely distributed data that has smaller variation.  The C-CASTERS algorithm provides the smallest mean and the shortest, thickest ``violins'' in all three shooter starting location categories. For all three categories, the C-CASTERS algorithm had more consistent results with smaller variance resulting in shorter ``violins''. Further, the quartiles were closer to the mean in the C-CASTERS performance than the other two algorithms indicating a more consistent performance over the 20 runs. Some themes common to all of the algorithms become apparent in Figure \ref{fig:WilsonSS}. The shooter starting in an exit is the worst case scenario for all of the algorithms and also has the greatest variance, longer "violins", meaning the number of casualties in each run can vary wildly. Further, a shooter starting in the hallway is the best case scenario for all of the algorithms and is also the most consistent with short, fat "violins". This is probably because in this case the algorithm benefits from the relatively higher confidence with which the shooter's movement can be predicted in the short term. The exits being centrally located makes the shooter's movement to either hallway equally likely and hence unpredictable. %While these results make sense for the Cyclic Dorm Floor, we now will examine if they remain consistent in the Acyclic School.

\begin{figure}[t]
\centering
\includegraphics[width=.99\columnwidth]{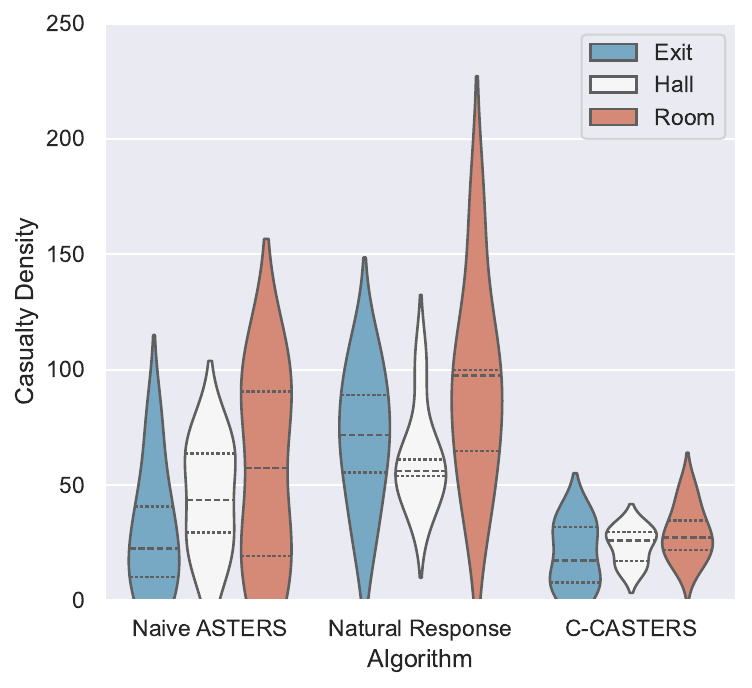}
\caption{Shooter's Spawn Location Effect - Acyclic School}
\label{fig:UnitySS}
\end{figure}

\textbf{Acyclic School Shooter's Start Location Effects:} The Acyclic School graphical environment performance results for the different shooter starting locations are displayed in Figure \ref{fig:UnitySS}. At first glance, it can be seen that the casualty distributions are less responsive to the shooter's start location in this graph compared to the cyclic graph. However, the C-CASTERS algorithm again performed best in each of the shooter start location categories, doing better than the other two algorithms in each situation. It may be worth remembering, that the Acyclic School environment contained almost twice as many evacuees as the Cyclic School Floor environment which explains the larger occurrences of casualties in the Acyclic School results. The C-CASTERS algorithm was the best performing of the three routing algorithms across the two environments and the different shooter starting location categories. 

It is very interesting to see that in the acyclic graph, no specific starting location could be identified that could qualify as ``most'' or ``least'' dangerous. The location of the three exits at the ends of the corridors assists in making predictable optimal choices for moving the evacuees. But at the same time, this illustrates the crucial role played by the capacity constrained optimization, since without that consideration, a large number of evacuees can be ``caught'' by the shooter near an exit. 

\subsection{Effect of Evacuee Distribution}

%We simulated two different distributions for the evacuees for each of the school environments. Our motivation for this was based on considerations of a normal school day where students change classrooms throughout the day. There will be times during the day when essentially no one will be in the hallways, while classes are in session for instance, and there will be times when the students are located in a mixture of both class rooms and hallways. The first distribution situation is called "Rooms" and the latter is called "Rooms and Halls". The number of evacuees spawned into the simulation is based upon the number of nodes classified in the evacuee distribution category. For instance, Cyclic Dorm Floor has hall nodes numbering from 1 to 29 and room nodes from 30 to 68. For a "Rooms" only distribution, the simulation would spawn four evacuees in each of the nodes from 30 to 68, while a "Rooms and Halls" distribution would result in four evacuees spawned in each node from 1 to 68.

As discussed in Section \ref{Sec:Evac_Dist}, to capture realistic scenarios in school evacuation planning, we modeled two distinct evacuee distribution scenarios for each simulated school environment. %This approach reflects the varying occupancy patterns observed during a typical school day, where the location of individuals fluctuates depending on class schedules and hallway usage. During class sessions, students are predominantly in classrooms with minimal hallway activity, whereas during transitions between classes, students occupy a combination of classrooms and hallways.
The two evacuee distribution scenarios are defined as follows:

% "Rooms" Distribution: Evacuees are located exclusively in classrooms.
% "Rooms and Halls" Distribution: Evacuees are distributed across both classrooms and hallways.
% Distribution Methodology

% The total number of evacuees in each scenario is determined by the number of nodes classified under the corresponding category. For example, in the Cyclic School Floor environment, nodes 1 through 29 represent hallway locations, while nodes 30 through 
% 68 correspond to classrooms.

\textit{Rooms Only} Distribution: In this configuration, evacuees are spawned exclusively in the classroom nodes (30 to 68), with four evacuees assigned to each node. This setup reflects a scenario where students are in their respective classrooms during a class session.
\textit{Rooms and Halls} Distribution: In this scenario, evacuees are uniformly distributed across all nodes (1 to 68), with four evacuees assigned to each node. This setup simulates a transition period when students are dispersed across both hallways and classrooms.

\begin{figure}[t]
\centering
\includegraphics[width=.98\columnwidth]{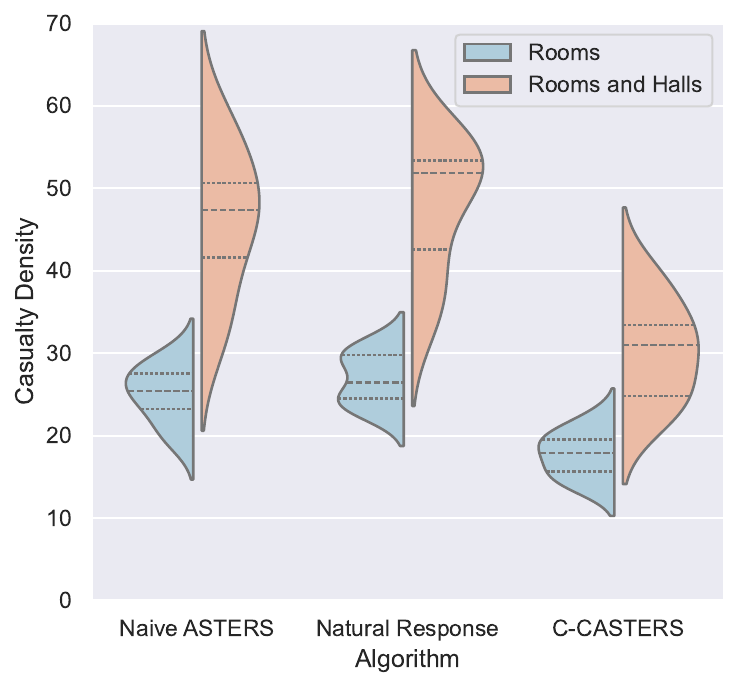}
\caption{Evacuee Distribution Effect - Cyclic School Floor}
\label{fig:WilsonED}
\end{figure}

Figure \ref{fig:WilsonED} presents split violin plots representing the casualty distributions for all parameter combinations within the Cyclic School Floor graphical environment, stratified by evacuee distribution scenarios. The left (blue) side of each violin plot corresponds to the ``Rooms Only'' distribution, while the right (red) side represents the ``Rooms and Halls'' distribution. These plots provide a comparative visualization of the casualty distributions for each routing algorithm under different initial evacuee placements.

\textbf{Performance Under the ``Rooms Only'' Distribution:} In the ``Rooms Only'' distribution scenario, all algorithms demonstrated relatively consistent performance. The casualty distributions for each algorithm are narrow and dense, with interquartile ranges closely clustered around their respective means. Among the algorithms, C-CASTERS achieved the best and most consistent results, with an average of $18.15$ casualties, Naive ASTERS followed with an average of $27.26$ casualties. Natural Response performed the least effectively, with an average of $28.31$ casualties.
The comparatively small casualty range in the ``Rooms Only'' distribution underscores the advantage of evacuees remaining in classrooms, where they are initially positioned out of the shooter’s line of sight. This scenario highlights the importance of minimizing unnecessary movement when evacuees are already in relatively safe locations.

\textbf{Performance Under the ``Rooms and Halls'' Distribution:} The ``Rooms and Halls'' distribution presented a stark contrast, with significantly larger casualty ranges and less consistent algorithmic performance. In this scenario, evacuees were dispersed across both classrooms and hallways, exposing them to greater risk due to proximity to the shooter’s potential path. The results for this distribution were as follows: C-CASTERS remained the best performer, with an average of $31.04$ casualties. Naive ASTERS had a higher average of $47.62$ casualties, while the Natural Response exhibited the poorest performance, with an average of $51.96$ casualties. The increased spread in the casualty data for all algorithms in the "Rooms and Halls" distribution indicates the variability and heightened difficulty of routing evacuees when hallway movement is involved.

The results depicted in Figure \ref{fig:WilsonED} emphasize the critical role of initial evacuee distribution in casualty outcomes during active shooter events. The superior performance of C-CASTERS, particularly under the challenging ``Rooms and Halls'' distribution, highlights its efficacy in reducing casualties even in less favorable conditions. These findings demonstrate that reducing movement through hallways during active shooter events is a critical goal for effective evacuation planning. Strategies should prioritize keeping evacuees in safe locations, such as classrooms, and only initiate movement when it provides a demonstrable improvement in safety. This principle is particularly crucial in the design and implementation of routing algorithms aimed at mitigating casualties in high-risk scenarios.

\begin{figure}[t]
\centering
\includegraphics[width=.98\columnwidth]{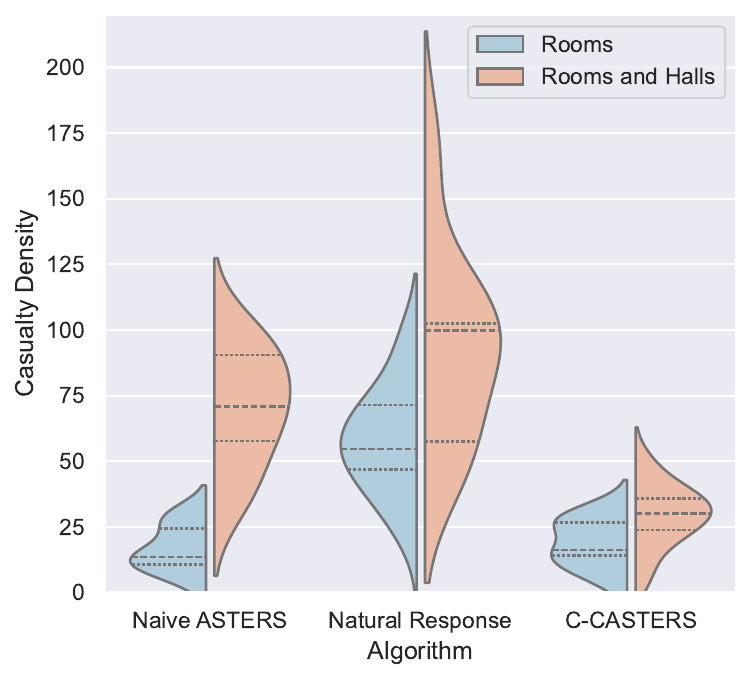}
\caption{Evacuee Distribution Effect - Acyclic School}
\label{fig:UnityED}
\end{figure}

Figure \ref{fig:UnityED} shows the casualty results from runs in the Acyclic School graphical environment. Again, the ``Rooms'' distribution runs showed a significant shift in the order of algorithm performance compared to the ``Rooms and Halls'' with greatly reduced mean casualties and variability among all of the algorithms. Naive ASTERS was the best and most consistent performer with an average casualty value of $13.78$, while the C-CASTERS followed closely achieving $15.53$ average casualties and the Natural Response producing $53.16$ average casualties in the ``Rooms'' distribution. The C-CASTERS algorithm significantly outperformed the other two algorithms in the ``Rooms and Halls'' case, with a very small and dense distribution of values. The C-CASTERS averaged $36.91$ casualties while the Naive ASTERS achieved $68.31$ average casualties and the Natural Response turned in $101.78$ average casualties. Again, across both graphical environments and evacuee distributions, the C-CASTERS algorithm was as good as or, more often, better than the other algorithms. The consistent theme across both graphical environments is the significantly better and more consistent outcomes produced when the evacuees start in the rooms regardless of algorithm. 

\begin{figure*}[t]
\centering
\includegraphics[width=1.98\columnwidth]{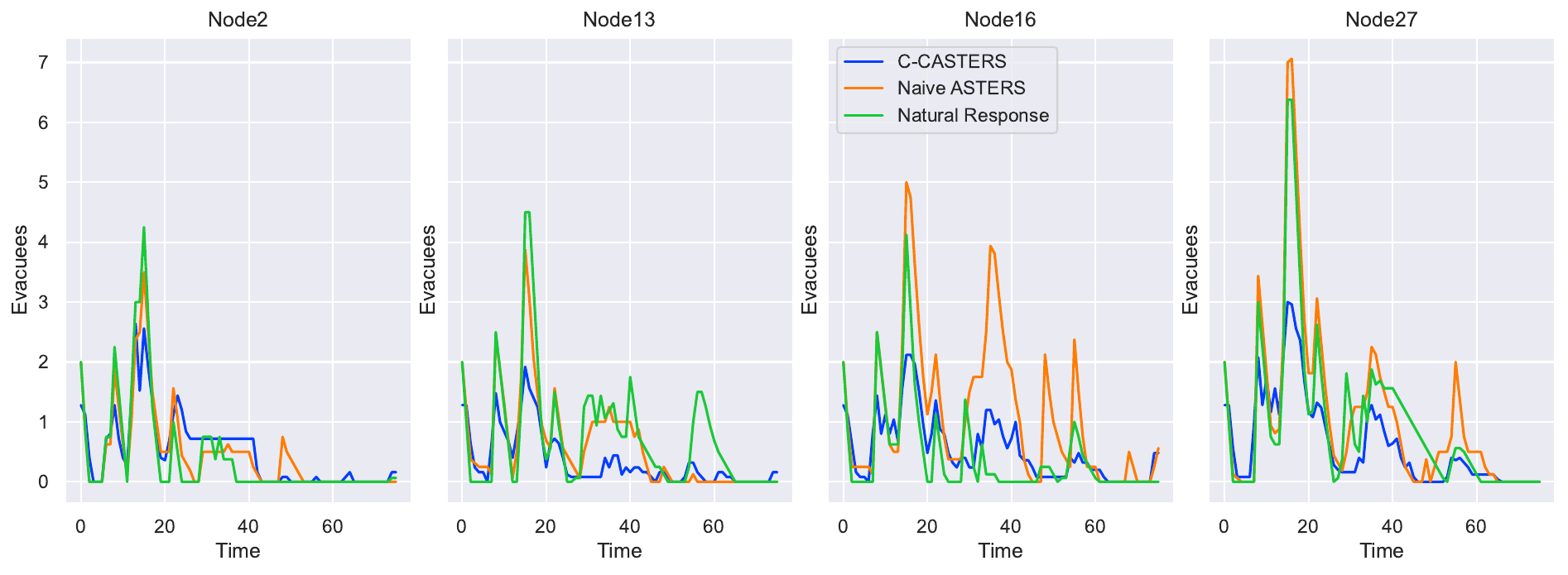}
\caption{Node Crowding Effect - Cyclic School Floor}
\label{fig:WilsonCrowd}
\end{figure*}

\subsection{Effect of Crowding}

%Our previous work was a starting point that was in essence a proof of concept that reinforcement learning was a promising approach to handling the movement of evacuees in an active shooter situation. However, we did not address the crowding of evacuees at exits or safe rooms as that would complicate the problem immensely while we were still exploring the approach. 
The effect of crowding and bottle-necking during an active shooter event is likely one of the most important elements related to the safety of the evacuees. The crowding of evacuees complicates the evacuation because it reduces the throughput of the affected nodes and slows down the movement of the evacuees. At the same time, it creates opportunities for potentially better routes to be used that avoid the larger crowds and would therefore be a faster means of exiting. In this paper we have explored a novel method for iteratively establishing routes for the evacuees and reserving their capacity along the route through time. Route reservation changes the  possible best routes for other nodes in future iterations. The order in which routes are created and distributed to the nodes is a design choice. In this case, we have prioritized nodes based on their distance from the exit which leads to the most efficient flow of evacuees. Below are the results of our node crowding analysis for each of the graphical environments.

%Figure \ref{fig:WilsonCrowd} shows the effect of crowding in the Cyclic Dorm Floor environment in the nodes of interest, which in this case are the nodes adjacent to the exits: $node_2$, $node_{13}$, $node_{16}$, and $node_{27}$. Each graph in the figure represents its average occupancy in number of evacuees at each instant of time from 0 seconds to 75 seconds which covers the time span when the majority of crowding occurs. Each graph for each node has a blue line (C-CASTERS), orange line (Naive ASTERS), and green line (Natural Response). Since we are looking at the effect of crowding and how it changes the ability of the evacuees to escape, we will examine the spikes in the graphs and not the more steady occupancy. Cyclic Dorm Floor contains fewer evacuees, so the crowding values are actually fairly small for each of the nodes, but, even with so few evacuees waiting, a few seconds can still mean the difference between life and death for them. During each spike in the node occupancy of the Cyclic Dorm Floor, the C-CASTERS achieves the minimum occupancy value among the routing algorithms in each and every one of the case spikes and often is approximately half the occupancy values of the other two algorithms. By iteratively identifying optimal routes for the nodes closest to the exits and farthest from shooter and then working farther from the exits and closer to the shooter, we are able to reduce the crowding in the key nodes thereby decreasing the number of casualties that occur and also explaining the better performance seen in the previous sections of the results. 

\textbf{Cyclic School Crowding Results:} Figure \ref{fig:WilsonCrowd} illustrates the impact of crowding in the Cyclic School Floor environment, with a focus on nodes adjacent to the exits: $node_2$, $node_{13}$, $node_{16}$, and $node_{27}$. The figure presents the average occupancy of each node, measured as the number of evacuees, over time from $t=0$ seconds to $t=75$ seconds. This time frame captures the critical period during which the majority of crowding occurs. Each graph contains three distinct curves representing different evacuation strategies: the blue line for C-CASTERS, the orange line for Naive ASTERS, and the green line for Natural Response.

To evaluate the effect of crowding on evacuee escape efficiency, we focus on spikes in node occupancy rather than steady-state values. Despite the relatively low number of evacuees in the Cyclic School Floor environment, crowding remains a significant factor. Even minor delays at exit-adjacent nodes can have life-threatening implications for evacuees.

The analysis reveals that, during each observed occupancy spike, C-CASTERS consistently achieves the lowest node occupancy values among the three routing algorithms. In many instances, its occupancy values are approximately half those of the Naive ASTERS and Natural Response algorithms. This superior performance is attributed to the iterative optimization strategy employed by C-CASTERS. By sequentially identifying optimal routes for nodes nearest the exits and farthest from the threat, and subsequently optimizing for nodes progressively farther from the exits and closer to the threat, the algorithm effectively reduces crowding at critical nodes. This reduction in crowding not only decreases casualties but also aligns with the improved performance metrics observed in preceding sections of the results.

The findings highlight the efficacy of C-CASTERS in mitigating the adverse effects of crowding during evacuation scenarios, thereby enhancing the safety and survival outcomes for evacuees.

\begin{figure*}[t!]
\centering
\includegraphics[width=1.98\columnwidth]{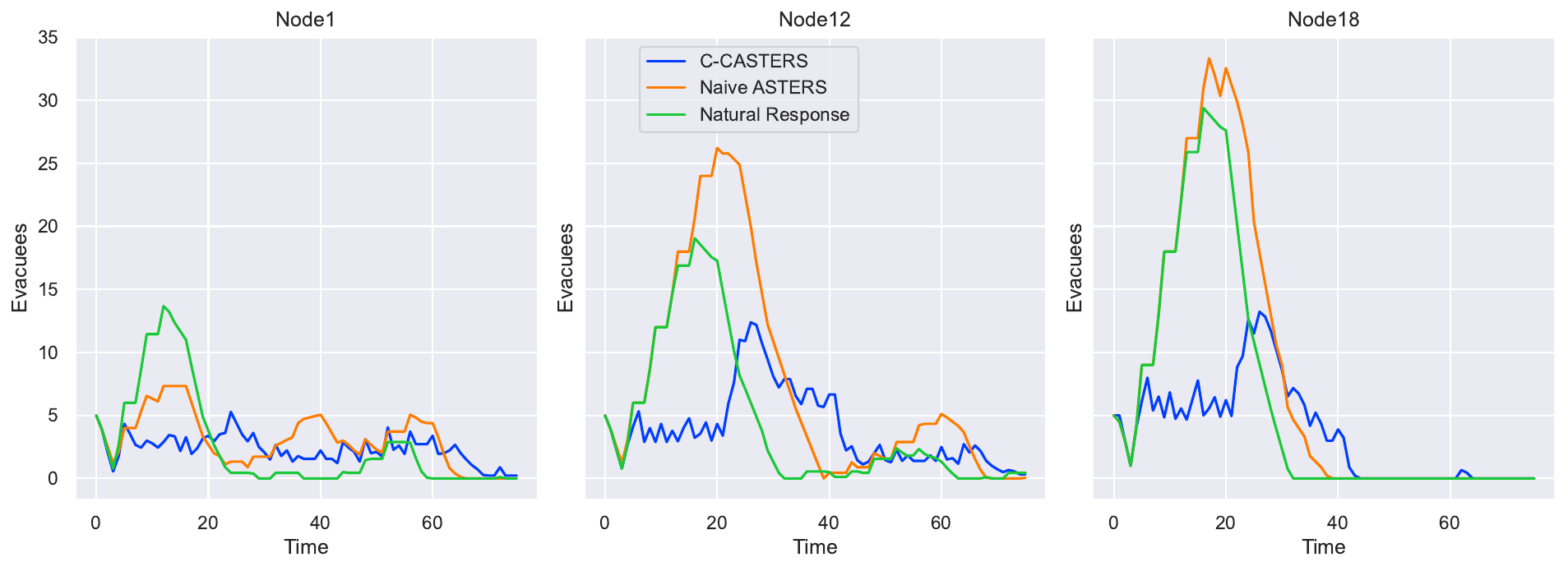}
\caption{Node Crowding Effect - Acyclic School}
\label{fig:UnityCrowd}
\end{figure*}

%Now, the Acyclic School has far more evacuees in it based upon the classroom sizes and the results for its simulations can be seen in Figure \ref{fig:UnityCrowd}. The nodes of interest for this graphical environment are also the nodes adjacent to the exits: $node_1$, $node_{12}$, and $node_{18}$. As you can see in the node graphs, the occupancy values are much larger in the spikes than in the Cyclic Dorm Floor environment. These spike, as large as they are, can cause up to 15 seconds of waiting to move through the exit. Again, as in the previous figure, the C-CASTERS always achieves the minimum occupancy value during each of the spike regions and is often less than half that of the other algorithms. It is interesting to note that the crowding spike doesn't even occur in $node_1$ for the C-CASTERS as it maintains a fairly constant occupancy preventing back ups. 

%Our approach to dealing with the issue of crowding appears to be effective in both of our graphical environments. While there is obvious improvements to optimize the process and no algorithm can completely prevent all crowding situations, our work shows promise in preventing the tragedies that have become all too common. 

\textbf{Acyclic School Crowding Results:} The Acyclic School environment, characterized by significantly larger classroom sizes, hosts a greater number of evacuees compared to the Cyclic School Floor environment. The results of the simulations for this environment are presented in Figure \ref{fig:UnityCrowd}. Similar to the previous analysis, the nodes of interest are those adjacent to the exits: $node_1$, $node_{12}$, and $node_{18}$.
In this environment, the occupancy spikes observed in the node graphs are notably larger than those in the Cyclic School Floor environment. These spikes can result in delays of up to 15 seconds as evacuees wait to pass through the exits. Despite the larger crowd sizes, the C-CASTERS algorithm consistently achieves the lowest occupancy values during these spikes, often maintaining values less than half those observed with the Naive ASTERS and Natural Response algorithms. Notably, $node_1$
exhibits no significant crowding spikes under C-CASTERS, as the algorithm effectively regulates occupancy to prevent congestion and maintain steady movement.

These results underscore the effectiveness of C-CASTERS in addressing crowding challenges. By proactively managing evacuation routes and iteratively optimizing for nodes closest to exits, the algorithm minimizes congestion and reduces delays, even under high-occupancy conditions.

While the results highlight the promise of this approach, further refinements are necessary to enhance its performance. No evacuation algorithm can fully eliminate crowding in all scenarios, but the findings presented here demonstrate the potential of C-CASTERS to mitigate critical crowding issues and reduce the likelihood of tragic outcomes during emergency evacuations.

\section{Conclusion and Future Work}

The study of capacitated transportation has far-reaching implications, and its relevance to emergency evacuations during active shooter scenarios cannot be overstated. Within this broader field, the focused investigation of routing individuals under such high-stakes circumstances demands sustained attention, innovation, and expansion. The importance of minimizing loss of life in these tragic events makes this an area of research worth pursuing. While we have addressed both non-capacitated and capacitated routing strategies, significant gaps remain in the literature, and meaningful solutions will require the collaborative efforts of a larger community of researchers.

Our research demonstrates the potential of reinforcement learning (RL) approaches to routing in active shooter situations. Specifically, the flexibility of C-CASTERS, which incorporates additional action options such as hiding in rooms or moving to safer nodes that are not necessarily exits, extends beyond the capabilities of traditional evacuation algorithms. Moreover, the algorithm’s iterative identification of optimal routes, coupled with capacity reservation along these routes, substantially reduces crowding at key nodes, thereby preventing unnecessary fatalities.

\subsection{Proposed Future Work}

Behavioral Adjustments of Evacuees: \\
Current simulations assume evacuees follow routing instructions without hesitation. In real-world scenarios, trauma and panic may hinder evacuees’ ability to act rationally or adhere to instructions. Modeling heterogeneous populations with varying probabilities of compliance, influenced by factors such as age, psychological state, and situational context, would add realism and complexity to evacuation models.

Integration of Deep Q-Learning (DQN) with Graph Neural Networks (GNNs): \\
Incorporating DQNs, which combine neural networks with Q-learning to handle high-dimensional state spaces, could facilitate real-time routing. Applying DQNs to GNNs, which are optimized for graph-based data, holds great potential for solving complex evacuation problems. Training these systems in simulated environments and deploying them in real-time scenarios could lead to a practical solution for routing in active shooter situations.

Generalization of Learning Across Environments: \\
To enhance the applicability of the algorithm, it is essential to generalize the learning process beyond specific training environments. By conducting offline training on diverse graphical environments and transitioning to online learning in new scenarios, researchers can evaluate the transferability of learned strategies. Using graph characteristics to identify similarities between environments could further improve the algorithm’s adaptability.

\subsection{Conclusion}
Our research highlights the potential of advanced routing algorithms like C-CASTERS in mitigating the devastating effects of active shooter incidents. While challenges remain, particularly in achieving real-time functionality, the proposed directions for future work aim to address these limitations and expand the scope of this critical research. By improving computational efficiency, incorporating behavioral realism, and exploring cutting-edge machine learning techniques, we can move closer to developing effective, scalable solutions for these emergencies.

The importance of this work lies in its ability to save lives, reduce casualties, and provide actionable solutions for individuals affected by active shooter events. We invite the broader research community to join this effort and contribute to the advancement of knowledge and technology in this essential domain.

\section*{Acknowledgement}
This material is based upon work supported by the U.S. Department of Homeland Security
under Grant Award 22STESE00001-03-02. The views and conclusions contained in this
document are those of the authors and should not be interpreted as necessarily representing
the official policies, either expressed or implied, of the U.S. Department of Homeland
Security.

\bibliographystyle{unsrt}
%\bibliography{trb_CAVIL}
\bibliography{2ndpaper, Ref_SC}

\end{document}